\documentclass{article} 
\usepackage{iclr2025_conference,times}

\usepackage{amsmath,amsfonts,bm}

\def\eqref#1{equation~\ref{#1}}

\def\1{\bm{1}}

\DeclareMathAlphabet{\mathsfit}{\encodingdefault}{\sfdefault}{m}{sl}
\SetMathAlphabet{\mathsfit}{bold}{\encodingdefault}{\sfdefault}{bx}{n}

\usepackage{hyperref}
\usepackage{url}

\usepackage{amsmath}
\usepackage{mathrsfs}
\usepackage{graphicx}
\usepackage{amsfonts}
\usepackage{verbatim}
\usepackage{subfigure}
\usepackage{booktabs,makecell,multirow}

\usepackage{float}
\usepackage{color}
\usepackage{bbm}
\usepackage[linesnumbered,ruled]{algorithm2e}
\usepackage{algorithmic}
\usepackage{multicol}
\usepackage{wrapfig}
\usepackage{bm}
\usepackage{array}
\usepackage{setspace}
\usepackage{enumitem}
\usepackage{soul}
\usepackage{minitoc}
\usepackage{caption}

\newcommand{\figureresolution}{high resolution}

\newcommand{\ourmodel}{VQ-CD}

\title{
Solving Continual Offline RL through Selective Weights Activation on Aligned Spaces
}

\author{%
  $\text{\textbf{Jifeng Hu}}^{\bm{1}}$~~~~$\text{\textbf{Sili Huang}}^{\bm{2}}$~~~~$\text{\textbf{Li Shen}}^{\bm{3}}$~~~~$\text{\textbf{Zhejian Yang}}^{\bm{4}}$
  $\text{\textbf{Shengchao Hu}}^{\bm{5}}$~~~~$\text{\textbf{Shisong Tang}}^{\bm{6}}$\\
  $\text{\textbf{Hechang Chen}}^{\bm{7}}\thanks{Corresponding Authors: chenhc@jlu.edu.cn (H. Chen) and lis221@lehigh.edu (L. Sun).}$~~~~$\text{\textbf{Yi Chang}}^{\bm{8}}$~~~~$\text{\textbf{Dacheng Tao}}^{\bm{9}}$~~~~$\text{\textbf{Lichao Sun}}^{\bm{10}*}$\\
  ${}^{\bm{1,2,4,7,8}}$School of Artificial Intelligence, Jilin University\\
  ${}^{\bm{3}}$School of Cyber Science and Technology, Shenzhen Campus of Sun Yat-sen University\\
  ${}^{\bm{5}}$MoE Key Lab of Artificial Intelligence, Shanghai Jiao Tong University\\
  ${}^{\bm{6}}$Shenzhen International Graduate School, Tsinghua University\\
  ${}^{\bm{9}}$College of Computing and Data Science, Nanyang Technological University\\
  ${}^{\bm{10}}$Computer Science and Engineering, Lehigh University\\
}

\iclrfinalcopy 
\begin{document}

\maketitle

\begin{abstract}

Continual offline reinforcement learning (CORL) has shown impressive ability in diffusion-based lifelong learning systems by modeling the joint distributions of trajectories.  
However, most research only focuses on limited continual task settings where the tasks have the same observation and action space, which deviates from the realistic demands of training agents in various environments.
In view of this, we propose Vector-Quantized Continual Diffuser, named \ourmodel{}, to break the barrier of different spaces between various tasks.
Specifically, our method contains two complementary sections, where the quantization spaces alignment provides a unified basis for the selective weights activation.
In the quantized spaces alignment, we leverage vector quantization to align the different state and action spaces of various tasks, facilitating continual training in the same space.
Then, we propose to leverage a unified diffusion model attached by the inverse dynamic model to master all tasks by selectively activating different weights according to the task-related sparse masks.
Finally, we conduct extensive experiments on 15 continual learning (CL) tasks, including conventional CL task settings (identical state and action spaces) and general CL task settings (various state and action spaces).
Compared with 16 baselines, our method reaches the SOTA performance.

\end{abstract}

\section{Introduction}

The endeavor of recovering high-performance policies from abundant offline samples gathered by various sources and continually mastering future tasks learning and previous knowledge maintaining gives birth to the issue of continual offline reinforcement learning (CORL)~\citep{levine2020offline, ada2024diffusion, huang2024solving}.
Ever-growing scenarios or offline datasets pose challenges for most continual RL methods that are trained on static data and are prone to showing catastrophic forgetting of previous knowledge and ineffective learning of new tasks~\citep{liu2024continual, zhang2023replay, korycki2021class}.
Facing these challenges, three categories of methods, rehearsal-based~\citep{huang2024solving, peng2023ideal, chaudhry2018efficient}, regularization-based~\citep{smith2023continual, zhang2023dynamics, zhang2022catastrophic}, and structure-based methods~\citep{zhang2023split, marouf2023weighted, borsos2020coresets}, are proposed to reduce forgetting and facilitate continual training.

However, most previous studies only focus on the continual learning (CL) setting with identical state and action spaces~\citep{liu2024continual, smith2023continual}.
It deviates from the fact that the ever-growing scenarios or offline datasets are likely to possess different state and action spaces with previous tasks for many reasons, such as the variation of demands and the number of sensors~\citep{yang2023continual, zhang2023split}.
Moreover, these datasets often come from multiple behavior policies, which pose the additional challenge of modeling the multimodal distribution of various tasks~\citep{ada2024diffusion, lee2024ad4rl}.
Benefiting from diffusion models' powerful expressive capabilities and highly competitive performance, an increasing number of researchers are considering incorporating them to address the CORL problems~\citep{ajay2022conditional, yue2024t, elsayed2024addressing} from the perspective of sequential modeling.
There have been several attempts to combine diffusion-based models with rehearsal-based and regularization-based techniques, which usually apply constraints to the continual model learning process with previous tasks' data or well-trained weights~\citep{smith2023continual, yue2024t, liu2024continual}.
However, constrained weight updating will limit the learning capability of new tasks and can not preserve the previously acquired knowledge perfectly~\citep{yang2023continual}. 
Although structure-based methods can eliminate forgetting and strengthen the learning capability by preserving well-trained weights of previous tasks and reserving disengaged weights for ongoing tasks, they are still limited in simple architecture and CL settings with identical state and action spaces~\citep{zhang2023split, wang2022dirichlet, mallya2018packnet}.
Thus, in this paper, we seek to answer the question:
\vspace{-5pt}
\begin{center}
\emph{Can we merge the merits of diffusion models' powerful expression and structure-based parameters isolation to master CORL problems with any task sequence?}
\end{center}
\vspace{-5pt}

We answer this in the affirmative through the key insight of allocating harmonious weights for each continual learning task.
Specifically, we propose Vector-Quantized Continual Diffuser called VQ-CD, which contains two complementary sections: the quantized spaces alignment (QSA) module and the selective weights activation diffuser (SWA) module.
To expand our method to any task sequences under the continual learning setting, we adopt the QSA module to align the different state and action spaces.
Concretely, we adopt vector quantization to map the task spaces to a unified space for training based on the contained codebook and recover it to the original task spaces for evaluation.
In the SWA module, we first perform task mask generation for each task, where the task masks are applied to the one-dimensional convolution kernel of the U-net structure diffusion model.
Then, we use the masked kernels to block the influence of unrelated weights during the training and inference.
Finally, after the training process, we propose the weights assembling to aggregate the task-related weights together for simplicity and efficiency.
In summary, our main contributions are fourfold:
\begin{itemize}[noitemsep,leftmargin=*]
\setlist{nolistsep} 
    \item We propose the Vector-Quantized Continual Diffuser (\ourmodel{}) framework, which can not only be applied to conventional continual tasks but also be suitable for any continual tasks setting, which makes it observably different from the previous CL method.
    \item In the quantized spaces alignment (QSA) module of \ourmodel{}, we adopt ensemble vector quantized encoders based on the constrained codebook because it can be expanded expediently. During the inference, we apply task-related decoders to recover the various observation and action spaces.
    \item In the selective weights activation (SWA) diffuser module of \ourmodel{}, we first perform task-related task masks, which will then be used to the kernel weights of the diffuser. After training, we propose assembling weights to merge all learned knowledge.
    \item Finally, we conduct extensive experiments on 15 CL tasks, including conventional CL settings and any CL task sequence settings. 
    The results show that our method surpasses or matches the SOTA performance compared with 16 representative baselines.
\end{itemize}

\section{Related Work}

\noindent\textbf{Offline RL.}~~~
Offline reinforcement learning is becoming an important direction in RL because it supports learning on large pre-collected datasets and avoids massive demand for expensive, risky interactions with the environments~\citep{mnih2015human, nair2020awac, cheng2022adversarially, ball2023efficient}.
Directly applying conventional RL methods in offline RL faces the challenge of distributional shift~\citep{ schaul2015prioritized, levine2020offline, xie2021policy, yue2022boosting} caused by the mismatch between the learned and data-collected policies, which will usually make the agent improperly estimate expectation return on out-of-distribution actions~\citep{schaul2015prioritized, kostrikov2021offline, ada2024diffusion}.
To tackle this challenge, previous studies try to avoid the influences of out-of-distribution actions by adopting constrained policy optimization~\citep{peng2019advantage, fujimoto2019off, nair2020awac, kostrikov2021offline}, behavior regularization~\citep{nachum2019algaedice, kumar2020conservative, ghosh2022offline}, importance sampling~\citep{jiang2016doubly, hallak2017consistent, zhang2020gendice}, uncertainty estimation~\citep{agarwal2020optimistic, wang2020qplex, lee2022offline}, and imitation learning~\citep{wang2020critic, siegel2020keep, chen2020bail}.

\noindent\textbf{Continual RL.}~~~
Continual learning (CL) aims to solve the plasticity
and stability trade-off under the task setting, where the agent can only learn to solve each task successively~\citep{zhang2023replay, wang2023distributionally}.
CL can be classified into task-aware CL and task-free CL according to whether there are explicit task boundaries~\citep{aljundi2019task, wang2023distributionally}.
In this paper, we mainly focus on task-aware CL.
There are three main technical routes to facilitate forward transfer (plasticity) and mitigate catastrophic forgetting (stability).
Rehearsal-based approaches~\citep{shin2017continual, mallya2018packnet, wang2022dirichlet, zhang2023split, smith2023continual} store a portion of samples from previous tasks and use interleaving updates between new tasks' samples and previous tasks' samples.
Simply storing samples increases the memory burden in many scenarios; thus, generative models such as diffusion models are introduced to mimic previous data distribution and generate synthetic replay for knowledge maintenance~\citep{zhai2019lifelong, qi2023better, gao2023ddgr}. 
Regularization-based approaches~\citep{kaplanis2019policy, kessler2020unclear, zhang2022catastrophic, zhang2023dynamics} seek to find a proficiency compromise between previous and new tasks by leveraging constraint terms on the total loss function.
Usually, additional terms of learning objectives will be adopted to penalize significant changes in the behaviors of models' outputs or the updating of models' parameters~\citep{kirkpatrick2017overcoming, kaplanis2019policy}.
In the structure-based approaches~\citep{wang2022dirichlet, kessler2022same, wang2022dirichlet, zhang2023split, smith2023continual, konishi2023parameter}, researchers usually consider parameter isolation by using sub-networks or task-related neurons to prevent forgetting.

\noindent\textbf{Diffusion RL.}~~~
Recently, diffusion-based models have shown huge potential in RL under the perspective of sequential modeling~\citep{sohl2015deep, ho2020denoising, rombach2022high, janner2022planning, ajay2022conditional, beeson2023balancing}.
A typical use of diffusion models is to mimic the joint distribution of states and actions, and we usually use state-action value functions as the classifier or class-free guidance when generating decisions~\citep{nichol2021improved, ho2022classifier, pearce2023imitating, liu2023more}.
Diffusion models, as representative generative models, can also be used as environmental dynamics to model and generate synthetic samples to improve sample efficiency or maintain previous knowledge in CL~\citep{yamaguchi2023limitation, hepburn2024model, lu2024synthetic, ding2024diffusion, liu2024continual}.
It is noted that the diffusion model's powerful expression ability on multimodal distribution also makes it suitable for being used as policies to model the distribution of actions and as planners to perform long-horizon planning~\citep{wang2022diffusion, kang2024efficient, chen2024simple}.
Besides, diffusion models can also be used as multi-task learning models to master several tasks simultaneously~\citep{he2024diffusion} or as multi-agent models to solve more complex RL scenarios~\citep{zhu2023madiff}.

\section{Preliminary}

\subsection{Continual Offline RL}

We focus on the task-aware CL in the continual offline RL in this paper~\citep{zhang2023replay, abel2023definition, wang2023distributionally, smith2023continual, qing2024advantage, wang2024critic}.
Suppose that we have $I$ successive tasks, and task $j$ arises behind task $i$ for any $i<j$.
Each task $i,i\in[1:I]$ is represented by a Markov Decision Process (MDP) $\mathcal{M}^i=\langle\mathcal{S}^i, \mathcal{A}^i, \mathcal{P}^i, \mathcal{R}^i, \gamma\rangle$, where we use supscript $i$ to differentiate different tasks, $I$ is the number of total tasks, $\mathcal{S}$ is the state space, $\mathcal{A}$ is the action space, respectively, $\mathcal{P}: \mathcal{S}\times\mathcal{A}\rightarrow \Delta(\mathcal{S})$ denotes the transition function, $\mathcal{R}: \mathcal{S}\times\mathcal{A}\times\mathcal{S}\rightarrow \mathbb{R}$ is the reward function, and $\gamma\in [0, 1)$ is the discount factor.
Conventional CL tasks have the same state and action spaces for all tasks, i.e., $|\mathcal{S}^i| = |\mathcal{S}^j|, |\mathcal{A}^i| = |\mathcal{A}^j|,\forall~i,j\in [1:I]$.
While for any tasks sequences, we have $|\mathcal{S}^i| \neq |\mathcal{S}^j|, |\mathcal{A}^i| \neq |\mathcal{A}^j|$.
In the offline RL setting, we can only access pre-collected datasets $\{D^i\}_{i\in [1:I]}$ from each task $\mathcal{M}^i$.
The goal of continual offline RL is to find an optimal policy that can maximize the discounted return $\sum_i^{I}\mathbb{E}_{\pi}[\sum_{t=0}^{\infty}\gamma^{t}r(s_{t}^{i}, a_{t}^{i})]$~\citep{fujimoto2021minimalist, yang2023continual, sun2023smart} on all tasks.

\subsection{Conditional Generative Behavior Modeling}

In this paper, we adopt a diffusion-based model with the U-net backbone as the generative model to fit the joint distribution $q(\tau_s)=\int q(\tau_s^{0:K}) d\tau_s^{1:K}$ of state sequences $\tau_s$ and an inverse dynamics model $\Psi(s_t, s_{t+1})$ to produce actions $a_t$, where $\tau_s$ is the tailored state sequences to facilitate training, $k\in [1:K]$ is the diffusion step, $t$ is the RL time step, and we omit the identification of tasks for the sake of simplicity because the training is same for all tasks.
Through specifying the pre-defined forward diffusion process $q(\tau_s^{k}|\tau_s^{k-1})=\mathcal{N}(\tau_s^{k};\sqrt{\alpha_k}\tau_s^{k-1},\beta_k\bm{I})$ and the trainable reverse process $p_{\theta}(\tau_s^{k-1}|\tau_s^{k})=\mathcal{N}(\tau_s^{k-1};\mu_{\theta}(\tau_s^{k},k), \Sigma^k)$~\citep{ho2020denoising}, we can train the diffusion model with the simplified loss function 
\begin{equation}\label{diffusion model train loss}
    \mathcal{L}(\theta)=\mathbb{E}_{k\sim U(1, 2, ..., K), \epsilon\sim\mathcal{N}(0,\bm{I}), \tau_s^0\sim D}[||\epsilon-\epsilon_{\theta}(\tau_s^k, k)||_2^2],
\end{equation}
where $\tau_s^k = \sqrt{\bar{\alpha}}\tau_s^0+\sqrt{1-\bar{\alpha}}\epsilon$, $\mu_{\theta}(\tau_s^{k})=\frac{1}{\sqrt{\alpha_k}}(\tau_s^k-\frac{\beta_k}{\sqrt{1-\bar{\alpha}_k}}\epsilon_{\theta}(\tau_s^{k},k))$, $\Sigma^k=\frac{1-\bar{\alpha}_{k-1}}{1-\bar{\alpha}_{k}}\beta_{k}\bm{I}$, $\alpha_{k}$ is the approximate discretization pre-defined parameters~\citep{chen2022offline, lu2023contrastive}, $\beta_k = 1 - \alpha_k$, $\bar{\alpha}_{k}=\prod_{\iota=1}^{k} {\alpha}_{\iota}$
, $U$ is the uniform distribution, $\epsilon$ is standard Gaussian noise, $\bm{I}$ is the identity matrix, we set the markov chain of SDE start from the sequential date of replay buffer $D$, i.e., $\tau_s^0=\tau_s$, and $\theta$ is the parameters of model $\epsilon_{\theta}$.

To achieve better control of action generation in RL, researchers propose classifier-guided and classifier-free guidance for diffusion models generation process $p_{\theta}(\tau_s^{k-1}|\tau_s^{k}, \mathcal{C})$~\citep{dhariwal2021diffusion, liu2023more}, where the conditions are $\mathcal{C}$, which is usually selected as Q function in RL.
Inspired by the Bayes expression $p_{\theta, \phi}(\tau_s^{k-1}|\tau_s^{k}, \mathcal{C})\propto p_{\theta}(\tau_s^{k-1}|\tau_s^{k})p_{\phi}(\mathcal{C}|\tau_s^{k})$ and the properties of exponential family distribution, classifier-guided methods propose to add guidance on the mean value with a separated training model, i.e., $p(\tau_s^{k-1}|\tau_s^{k}, \mathcal{C})=\mathcal{N}(\mu_{\theta}+\Sigma^k \cdot \nabla log~p_{\phi}(\mathcal{C}|\tau_s), \Sigma^k)$.
Classifier-free methods propose to model the correlation between the state sequences and conditions in the training phase by learning unconditional and conditional noise $\epsilon_{\theta}(\tau_s^k, \emptyset, k)$ and $\epsilon_{\theta}(\tau_s^k, \mathcal{C}, k)$, where $\emptyset$ is usually the zero vector~\citep{ajay2022conditional}.
At each diffusion step $k$, the perturbed noise is calculated by $\epsilon_{\theta}(\tau_s^k, \emptyset, k)+\omega(\epsilon_{\theta}(\tau_s^k, \mathcal{C}, k)-\epsilon_{\theta}(\tau_s^k, \emptyset, k))$.

\section{Method}
\begin{figure*}[t!]
 \begin{center}
\includegraphics[angle=0,width=0.99\textwidth]{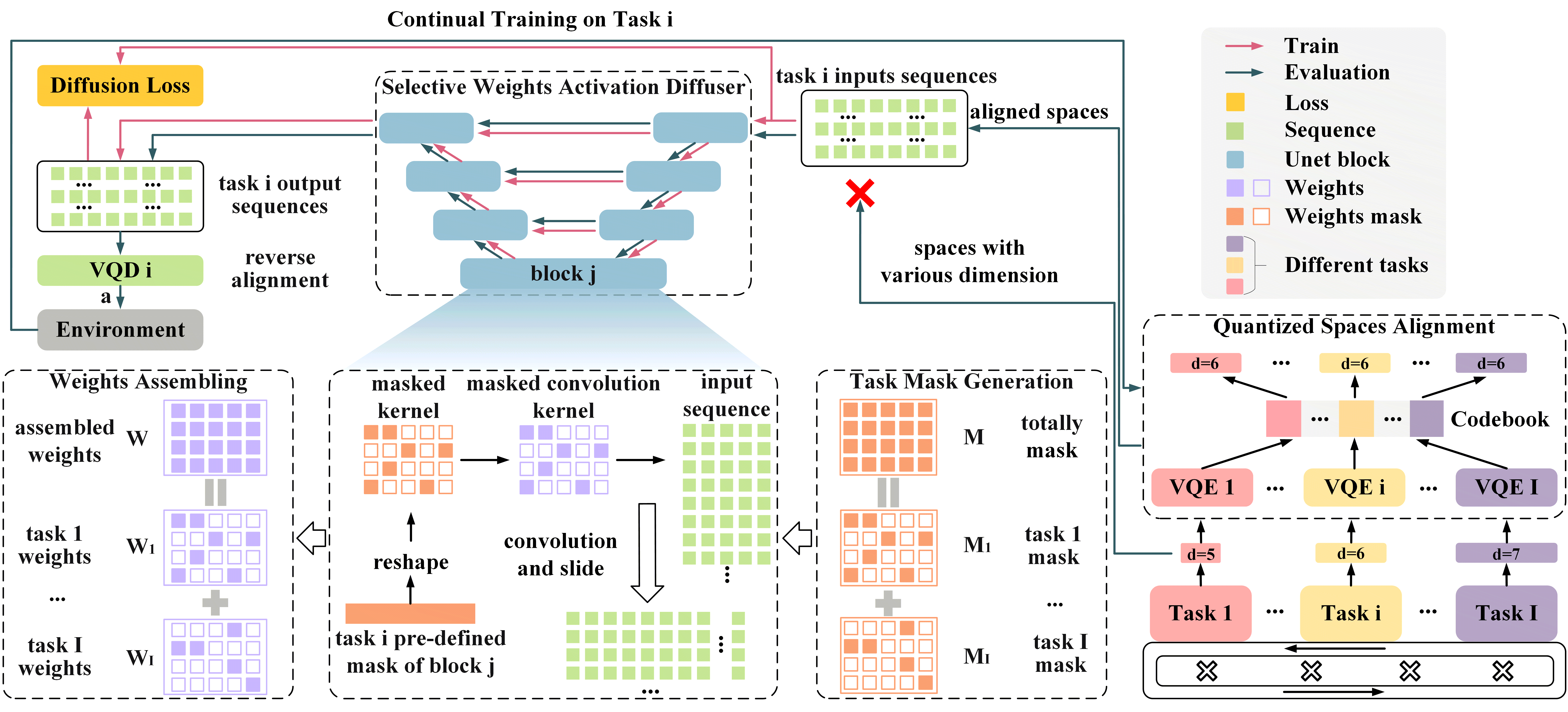}
 \caption{The framework of \ourmodel{}. It contains two sections: The Quantized Space Alignment (QSA) module and the Selective Weights Activation (SWA) module, where QSA enables our method to adapt for any continual learning task setting by transferring the different state and action spaces to the same spaces. SWA uses selective neural network weight activation to maintain the knowledge of previous tasks through task-related weight masks. After the training, we perform weights assembling to integrate the total weights and save the memory budget.}
 \label{framework}
 \end{center}
 \vspace{-0.3cm}
 \end{figure*}

Our method enables training on any CL task sequences through two sections (as shown in Figure~\ref{framework}): the selective weights activation diffuser (SWA) module and the quantized spaces alignment (QSA) module.
The detailed algorithm is shown in Algorithm~\ref{algorithm} of Appendix~\ref{appendix of pseudocode}.
In the following parts, we introduce these two modules in detail.

\subsection{Quantized Spaces Alignment}
To make our method suitable for solving any CL task sequence setting, we propose aligning the different state and action spaces with the quantization technique.
Specifically, we propose to solve the following quantized representation learning problem
\begin{equation}
\label{QSA problem}
\begin{aligned}
& \underset{\theta_e,\theta_d,\theta_q}{\text{min}}
& & \mathcal{L}_{QSA}(x;\theta_e,\theta_d,\theta_q), \\
& \text{s.t.}
& & ||z_q||_2^2<\rho,
\end{aligned}
\end{equation}
where $\mathcal{L}_{QSA}(x)=\mathbb{E}\left[||x-f_{VQD}(z_q;\theta_d)||_2^2\right]+\mathbb{E}\left[||\mathbf{sg}(z_q)-z_e||_2^2\right]+\mathbb{E}\left[||\mathbf{sg}(z_e)-z_q||_2^2\right]$ is the total quantized loss, $\mathbf{sg}(\cdot)$ represents the stop gradient operation, $\theta_e$ and $\theta_d$ are the parameters of the vector quantized encoder (VQE) and vector quantized decoder (VQD), $\theta_q$ is the parameters of the codebook, $\rho$ limits the range of codebook embeddings, $x$ can represent the states or actions for each specific CL task, $z_q = f_{\theta_q}(z_e)$ is the quantized representation which is consisted of fixed number of fixed-length quantized vectors, and $z_e = f_{VQE}(x;\theta_e)$ is the output of the encoder.
Here, we propose searching the constrained optimal solution of the above problem for the consideration of the diffusion model training within a limited value range, just like the limit normalization in CV~\citep{ho2020denoising, dhariwal2021diffusion} and RL~\citep{ajay2022conditional, lu2023contrastive}.
There are many methods to force optimization under restricted constraints, such as converting the constraints to a penalty term~\citep{boyd2004convex}.
In our method, for simplicity and convenience, we propose to directly clip the quantized vector $z_q$ to meet the constraints after every codebook updating step. 
Moreover, to meet the potential demand for extra tasks beyond the predefined CL tasks, we design the codebook as easy to equip, where the quantized spaces of different tasks are separated so that we can expediently train new task-related encoders, decoders, and quantized vectors.

For tasks where the state and action spaces are different, we can use the well-trained QSA module to obtain the aligned state feature $s^i_{z_q}=f^i_{s,\theta_q}(f^i_{VQE_s}(s^i;\theta_e))$ and the action feature $a^i_{z_q}=f^i_{a,\theta_q}(f^i_{VQE_a}(a^i;\theta_e))$ for each task $i$.
Thus, we can use $\tau^i_{s_{z_q}}$ and $\tau^i_{a_{z_q}}$ to represent the state and action feature sequences.

\subsection{Selective Weights Activation}

In this section, we introduce how to selectively activate different parameters of the diffusion model to reduce catastrophic forgetting and reserve disengaged weights for ongoing tasks.

\noindent\textbf{Task Mask Generation.}~~
Suppose that the diffusion model contains $L$ blocks, and the weights (i.e., parameters) of block $l$ are denoted by $W_l,l\in\{1,...,L\}$.
There are two ways to disable the influence of the weights on the model outputs.
One is masking the output neurons $O_l=f_l(\cdot;W_l)$ of each block, where $f_{l}(\cdot)$ is the neural network function of block $l$.
This strategy is friendly to MLP-based neural networks for two reasons: 1) the matrix calculation, such as $W_l*x$, is relatively simple so that we can easily recognize the disabled weights; 2) we do not need to apply any special operation on the optimizer because the output masking will cut off gradient flow naturally.
However, we can not arbitrarily apply the above masking strategy to more expressive network structures, such as convolution-based networks, because we can not easily distinguish the dependency between parameters and outputs.
Thus, we search for another masking strategy: masking the parameters $W_l$ with $M_l$, which permits us to control each parameter accurately.

Specifically, suppose that the total available mask positions of block $l$ are $M_l$. 
In this paper, $M_l$ is a ones matrix, and the entries with 0 mean that we will perform masking.
Before training on task $i$, we first pre-define the specific mask $M_{i,l}$ of task $i$ on block $l$ by randomly sampling unmasked positions from the remaining available mask positions.
Then, with the increase of the tasks, the remaining available mask positions decrease until $M_l=\sum_{i=1}^{I} M_{i,l}$.

\noindent\textbf{Selective Weights Forward and Backward Propagation.}~~
After obtaining the mask $M_{i,l}$, we can perform forward propagation with masked weights
\begin{equation}\label{forward propagation masking}
\begin{aligned}
& \epsilon_{\theta}(\tau_{\cdot}^k, k) = f_L(f_{L-1}(...(f_1(\cdot))))\\
& O_{l+1} = f_{l+1}(O_{l}, k;M_{i,l+1}\circ W_{l+1}), O_{0}=\tau^{i,k}_{s_{z_q}}/\tau^{i,k}_{a_{z_q}},
\end{aligned}
\end{equation}
where $\epsilon_{\theta}$ is the noise prediction model introduced in Equation~\ref{diffusion model train loss}, and $M_{i,l+1}\circ W_{l+1}$ represents the pairwise product. 
$\tau^{i,k}_{s_{z_q}}$ and $\tau^{i,k}_{a_{z_q}}$ denote the perturbed state or action sequences of task $i$ at diffusion step $k$.
Through forward designing, we can selectively activate different weights for different tasks through the mask $M_{i,l+1}$, thus preserving previously acquired knowledge and reserving disengaged weights for other tasks. 
Though we can expediently calculate the masked output $O_{l+1}$ during forward propagation with weights or neurons masking, it poses a challenge to distinguishing the dependency from weights to loss and updating the corresponding weights during the backward propagation.
In order to update the corresponding weights, we realize two methods. 
1) Intuitively, we propose to update the neural network with the sparse optimizer rather than the dense optimizer~\cite{diederik2014adam}, where the position and values of the parameters are recorded to update the corresponding weights. 
However, in the implementation, we find that the physical time consumption of the sparse optimizer is intractable (Refer to Table~\ref{The optimizer time consumption} of Appendix~\ref{Time Consumption of Different Optimizers} for more details.), which encourages us to find a more straightforward and convenient method.
2) Thus, we propose extracting and assembling the corresponding weights at the end of the training rather than updating the corresponding weights during training.
This choice brings two benefits: (1) It can significantly reduce the time consumption spent on training. (2) It is friendly to implementation on complex network structures.

\noindent\textbf{Weights Assembling.}~~
Assembling weights after training permits us to save the total acquired knowledge and do not need extra memory budgets.
Concretely, after training on task $i$, we will obtain the weights $W_{i}$, which can be extracted with the mask $M_i$ from the total weights $W[i*\Omega]$, including all the diffusion model weights.
We use $W_{i}$ to denote the weights related to task $i$, $\Omega$ is the training step on each CL task, and $W[i*\Omega]$ represents the total weight checkpoint at training step $i*\Omega$.
Then, at the end of the training, we can assemble weights $\{W_{i}|i\in I\}$ by simply adding these weights together because of the exclusiveness property, i.e., $W=\sum_{i=1}^{I}W_i=\sum_{i=1}^{I}M_i\circ W[i*\Omega]$.

\section{Experiments}

In this section, we will introduce environmental settings, evaluation metrics, and baselines in the following sections.
Then, we will report and analyze the comparison results, ablation study, and parameter sensitivity analysis.
Other implementation details are shown in Appendix~\ref{hyperparameters} and ~\ref{computation}.

\subsection{Environmental Settings}\label{Environmental Settings}
Following previous studies~\citep{zhang2023replay, yang2023continual}, we select MuJoCo Ant-dir and Continual World (CW) to formulate traditional CL settings with the same state and action spaces.
In Ant-dir, we select several tasks, such as 10-15-19-25 and 4-18-26-34-42-49, for training and evaluation.
In CW, we adopt the task setting of CW10, which contains 10 robotic manipulation tasks for CL performance comparison.
Additionally, we propose to leverage D4RL tasks~\citep{fu2020d4rl} to construct the CL settings with diverse state and action spaces.
We select the Hopper, Walker2d, and HalfCheetah as elements to construct CL tasks, where each environment among Hopper, Walker2d, and HalfCheetah contains 6 qualities (random, medium, expert, medium-expert, medium-replay, and full-replay) datasets.

\subsection{Evaluation Metrics}\label{Evaluation Metrics}

Considering the various reward structures of different environments, we should adopt different performance comparison metrics.
For Ant-dir, we adopt the average episodic return over all tasks as the performance comparison, i.e., the final performance $P=\text{mean}(\sum_i R_i)$ is calculated based on the task $i$'s return $R_i$.
In the CW environment, previous works~\citep{wolczyk2021continual, anand2023prediction} usually adopt the success rate $\Psi$ as the performance metric.
Thus, we adopt the average success rate on all tasks as the final performance, i.e., $P=\text{mean}(\sum_i \Psi_i)$.
For the D4RL environments, we use the normalized score $\Phi$~\citep{wang2022diffusion, huang2024solving} as the metric to calculate the performance $P=\text{mean}(\sum_i \Phi_i)$, where $\Phi_i=\frac{R_i-R_{random}}{R_{expert}-R_{random}}*100$.
Usually, we can use the interface of these environments to obtain the score expediently.

\subsection{Baselines}\label{Baselines}
We select various representative CL baselines, which can be classified into diffusion-based and non-diffusion-based methods.
For example, the diffusion-based methods consist of CRIL~\citep{gao2021cril}, DGR~\citep{shin2017continual}, t-DGR~\citep{yue2024t}, MTDIFF~\citep{he2023diffusion}, CuGRO~\citep{liu2024continual}, CoD~\citep{hu2024continual}, and CoD variants.
The non-diffusion-based methods include EWC~\citep{kirkpatrick2017overcoming}, PackNet~\citep{mallya2018packnet}, Finetune, IL-rehearsal~\citep{wan2024lotus}, and Multitask.
From the perspective of mainstream CL classification standards, these baselines can also be sorted as rehearsal-based methods (CRIL, DGR, t-DGR, CoD, and IL-rehearsal), regularization-based methods (EWC, CuGRO, and Finetune), and structure-based methods (PackNet, Multitask, and MTDIFF).

 \begin{figure*}[t!]
 \begin{center}
\ifthenelse{\equal{\figureresolution}{low resolution}}
    {\includegraphics[angle=0,width=0.99\textwidth]{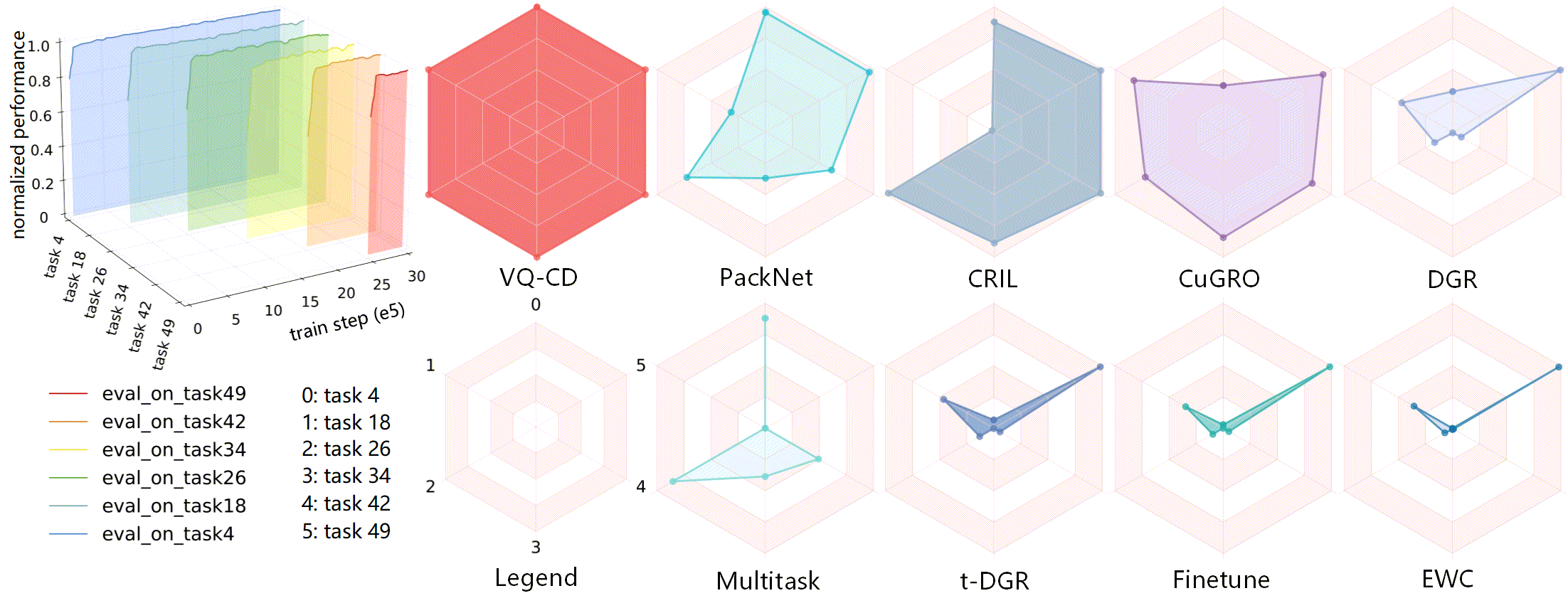}}
    {\includegraphics[angle=0,width=0.99\textwidth]{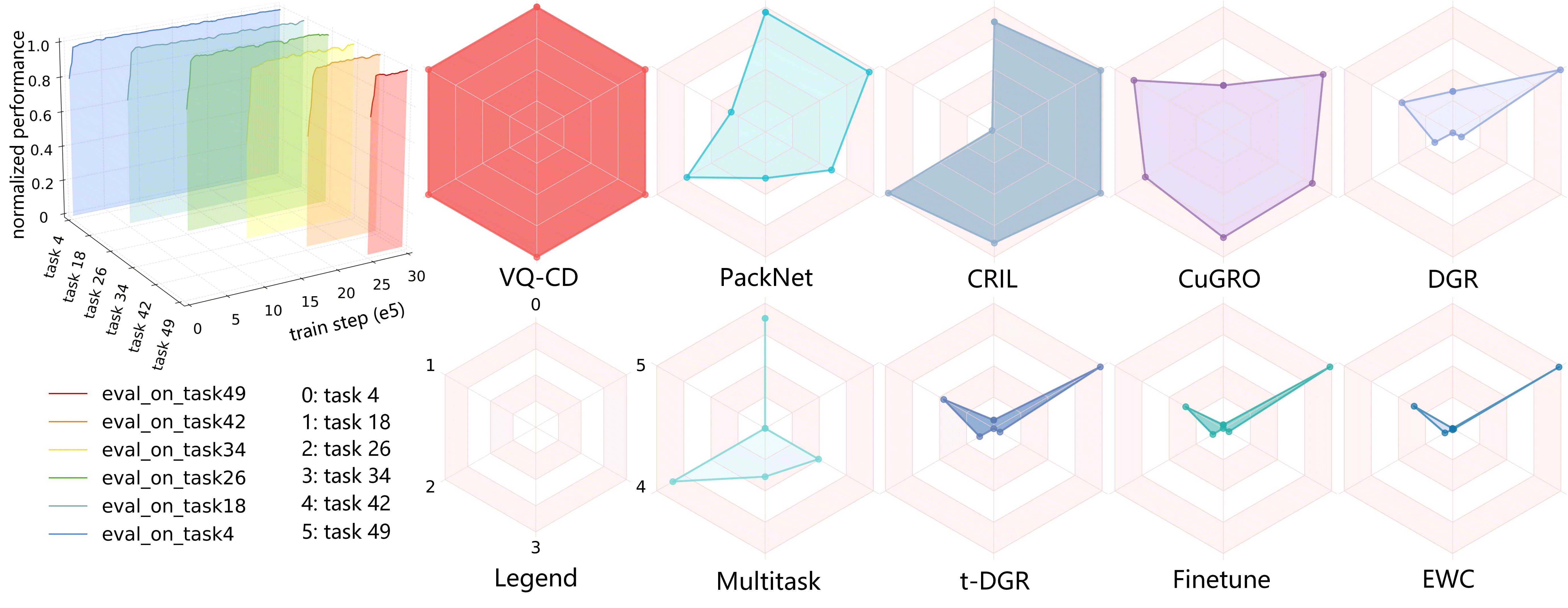}}
 \caption{The comparison of \ourmodel{} and several baselines on the continual tasks setting (Ant-dir task 4-18-26-34-42-49). We train on each task for 500k steps. We report the normalized evaluation performance of \ourmodel{} in the top left corner, where the coordinates, e.g., task 4, represent evaluation on task 4 at different training tasks. To show the overall performance on all tasks, we show the normalized evaluation performance on the six tasks after finishing the training at the right part.}
 \end{center}
 \end{figure*}

\begin{table}[t!]
\centering
\vspace{-1em}
\small
\caption{The comparison of \ourmodel{}, diffusion-based baselines, and LoRA methods on Ant-dir tasks, where the continual task sequence is 10-15-19-25. The results are average on 30 evaluation rollouts with 30 random seeds.}
\label{Ant-dir compare with diffusion-based models}
\resizebox{0.99\textwidth}{!}{
\begin{tabular}{l | r | r | r | r | r | r | r | r | r}
\toprule
\specialrule{0em}{1.5pt}{1.5pt}
\toprule
Method & \makecell[r]{\ourmodel{}\\(\textbf{ours})} & CoD & \makecell[r]{Multitask \\ CoD} & \makecell[r]{IL-\\rehearsal} & \makecell[r]{CoD-\\LoRA} & \makecell[r]{Diffuser-w/o \\ rehearsal} & \makecell[r]{CoD-\\RCR} & MTDIFF & \makecell[r]{DD-w/o \\ rehearsal} \\
\midrule[1pt]
\makecell[l]{Mean\\return} & 558.22\tiny{$\pm$1.14} & 478.19\tiny{$\pm$15.84} & 485.15\tiny{$\pm$5.86} & 402.53\tiny{$\pm$17.67} & 296.03\tiny{$\pm$11.95} & 270.44\tiny{$\pm$5.54} & 140.44\tiny{$\pm$32.11} & 84.01\tiny{$\pm$41.10} & -11.15\tiny{$\pm$45.27} \\
\bottomrule
\specialrule{0em}{1.5pt}{1.5pt}
\bottomrule
\end{tabular}}
\vspace{-0.3cm}
\end{table}

\subsection{Experimental Results}\label{Experimental Results}

In this section, we mainly separate the experimental settings into two categories, the traditional CL settings with the same state and action spaces and the arbitrary CL settings with different state and action spaces, to show the effectiveness of our method.
Besides, we also investigate the influence of the alignment techniques, such as auto-encoder, variational auto-encoder, vector-quantized variational auto-encoder (we adopt this in our method).
More deeply, we investigate how to deal with the potential demand for additional tasks beyond the pre-defined task length by releasing nonsignificant masks or expanding more available weights (Refer to Appendix~\ref{Supporting Tasks Training Beyond the Pre-defined Task Sequence} for more details.).

The traditional CL settings correspond to the first question we want to answer: \emph{Can \ourmodel{} achieve superior performance compared with previous methods in the traditional CL tasks?}

\begin{figure*}[t!]
 \begin{center}
\ifthenelse{\equal{\figureresolution}{low resolution}}
    {\includegraphics[angle=0,width=0.99\textwidth]{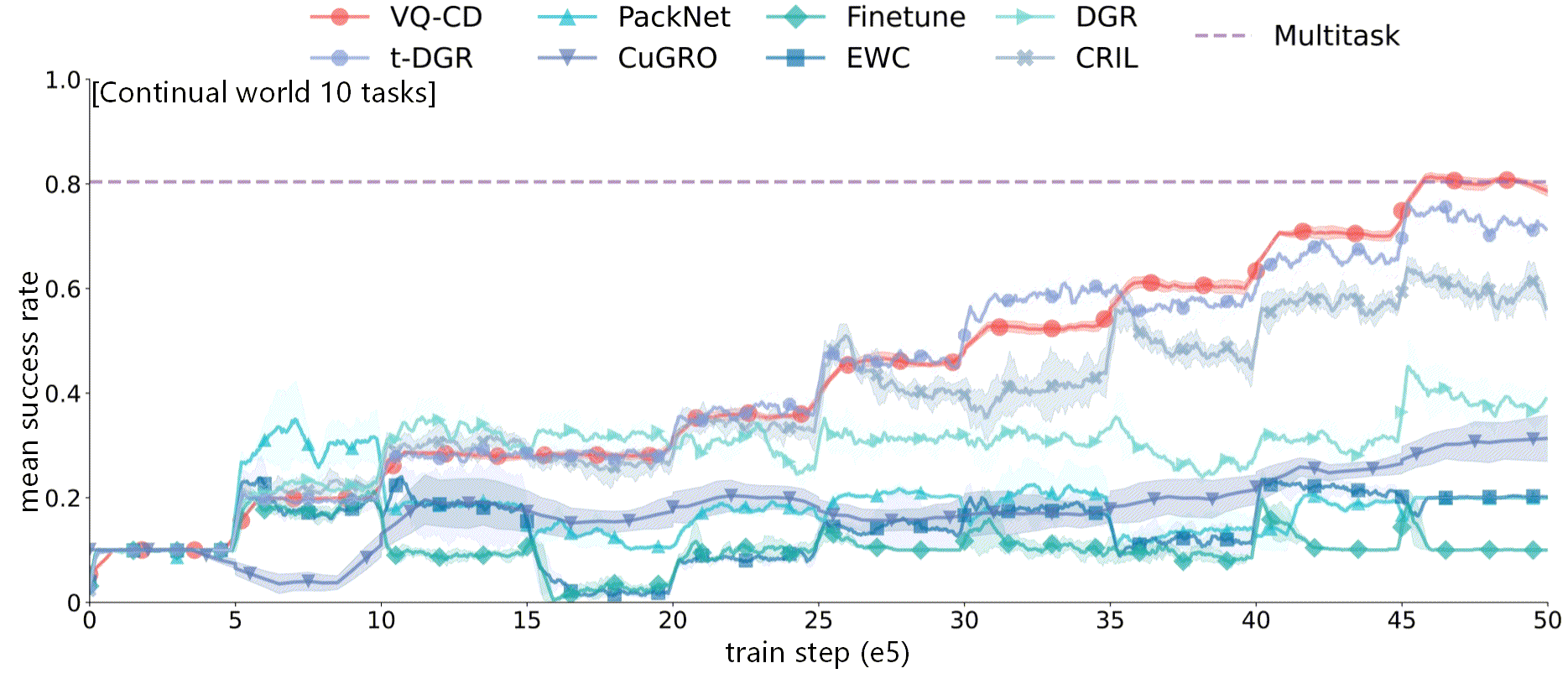}}
    {\includegraphics[angle=0,width=0.99\textwidth]{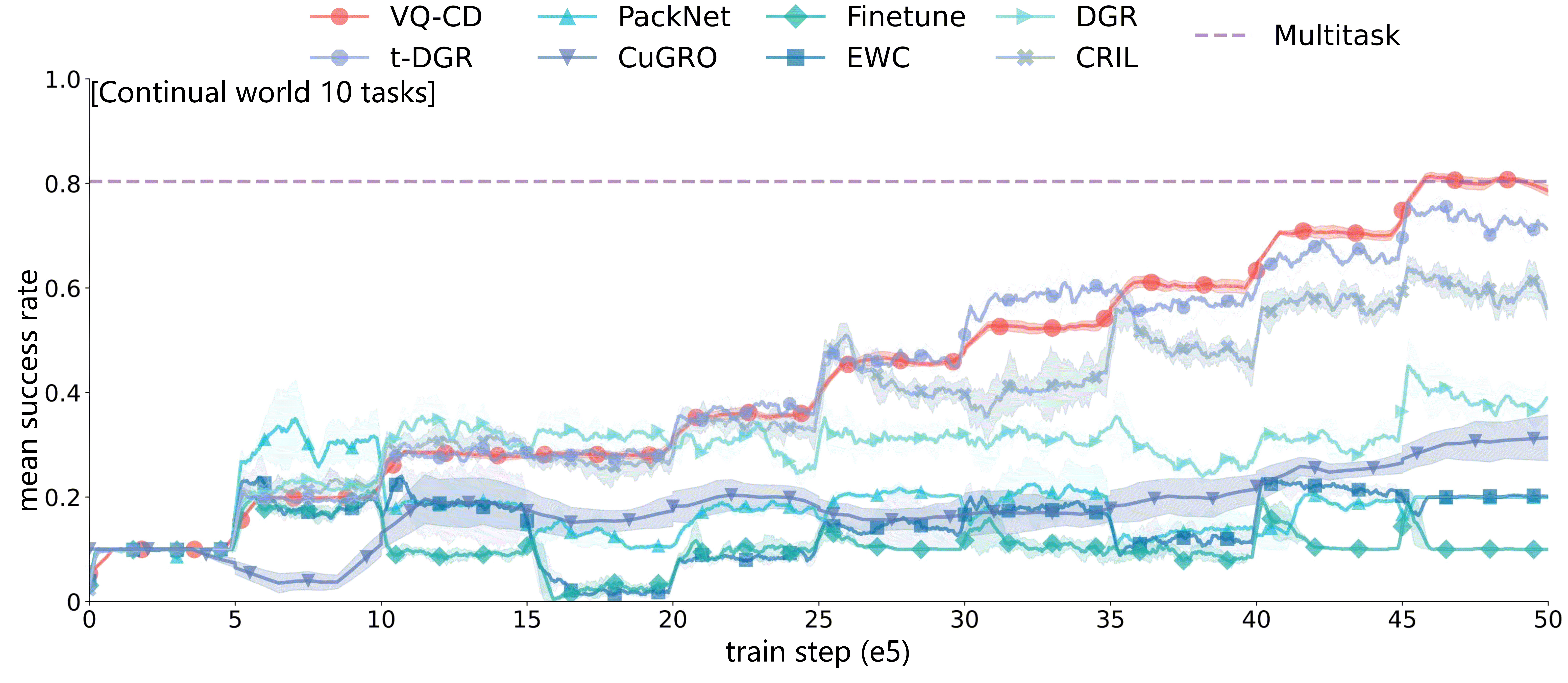}}
 \caption{The experiments on the CW10 tasks, which contain various robotics control tasks. We train each method on each task for 5e5 steps and use the mean success rate on all tasks as the performance metric. Generally, we can see the superiority of our method from the above figure.}
 \label{CW10 compare with classical CL methods}
 \end{center}
 \end{figure*}

We use Ant-dir and Continual World~\citep{zhang2023replay, yang2023continual} to formulate the continual task sequence, where we select ``10-15-19-25'' as the task sequence in Ant-dir and ``hammer-v2, push-wall-v2, faucet-close-v2, push-back-v2, stick-pull-v2, handle-press-side-v2, push-v2, shelf-place-v2, window-close-v2, peg-unplug-side-v2'' to construct CW10 CL setting.
For simplicity, we do not align the state and action spaces with quantized alignment techniques because the traditional CL setting naturally has the same spaces.
The comparison results between our method and several diffusion-based baselines are shown in Table~\ref{Ant-dir compare with diffusion-based models}, where these baselines include rehearsal-based (CoD and Il-rehearsal), parameter-sharing (CoD-LoRA), multitask training (Multitask CoD and MTDIFF), and representative diffusion RL methods (Diffuser-w/o rehearsal, CoD-RCR, and DD-w/o rehearsal).
Our method surpasses all baselines in the Ant-dir task setting by a large margin, which directly shows the effectiveness of our method.
As another experiment of CL setting with the same state and action spaces, we report the results in Figure~\ref{CW10 compare with classical CL methods}.
Compared with the upper bound performance of Multitask, our method reaches the same performance after the CL training.
With the increasing of new tasks, our method continually masters new tasks and sustains the performance while the baselines show varying degrees of performance attenuation, which can be found in the fluctuation of the curves.
Moreover, the final performance difference between one method and the Multitask method indicates the forgetting character, which can be reflected by the overall upward trend of these curves.
More experiments of shuffling task orders can be found in Appendix~\ref{Experiments of Task Order Shuffling}.

 \begin{figure*}[t!]
\vspace{-1em}
 \begin{center}
\ifthenelse{\equal{\figureresolution}{low resolution}}
    {\includegraphics[angle=0,width=0.99\textwidth]{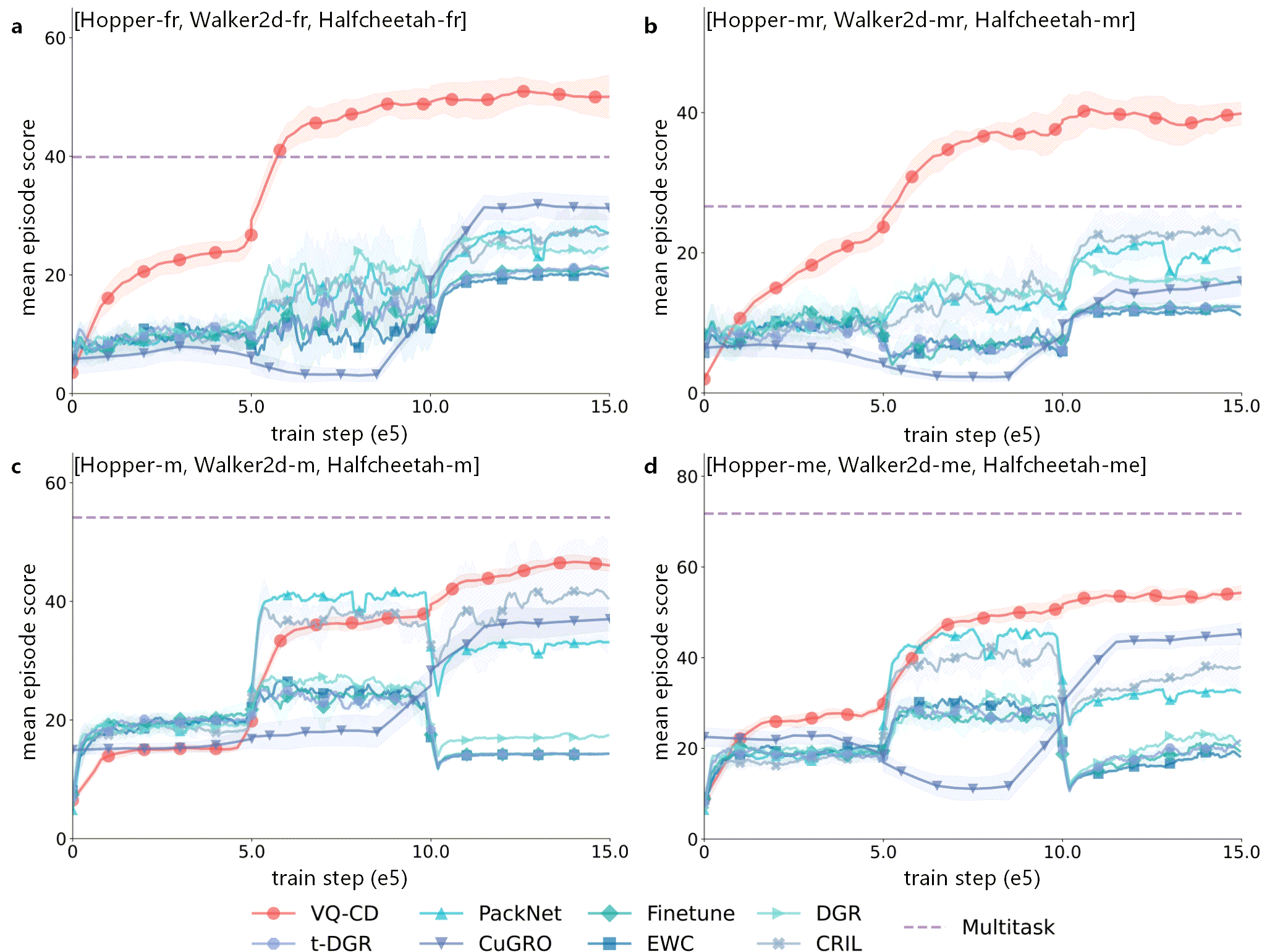}}
    {\includegraphics[angle=0,width=0.99\textwidth]{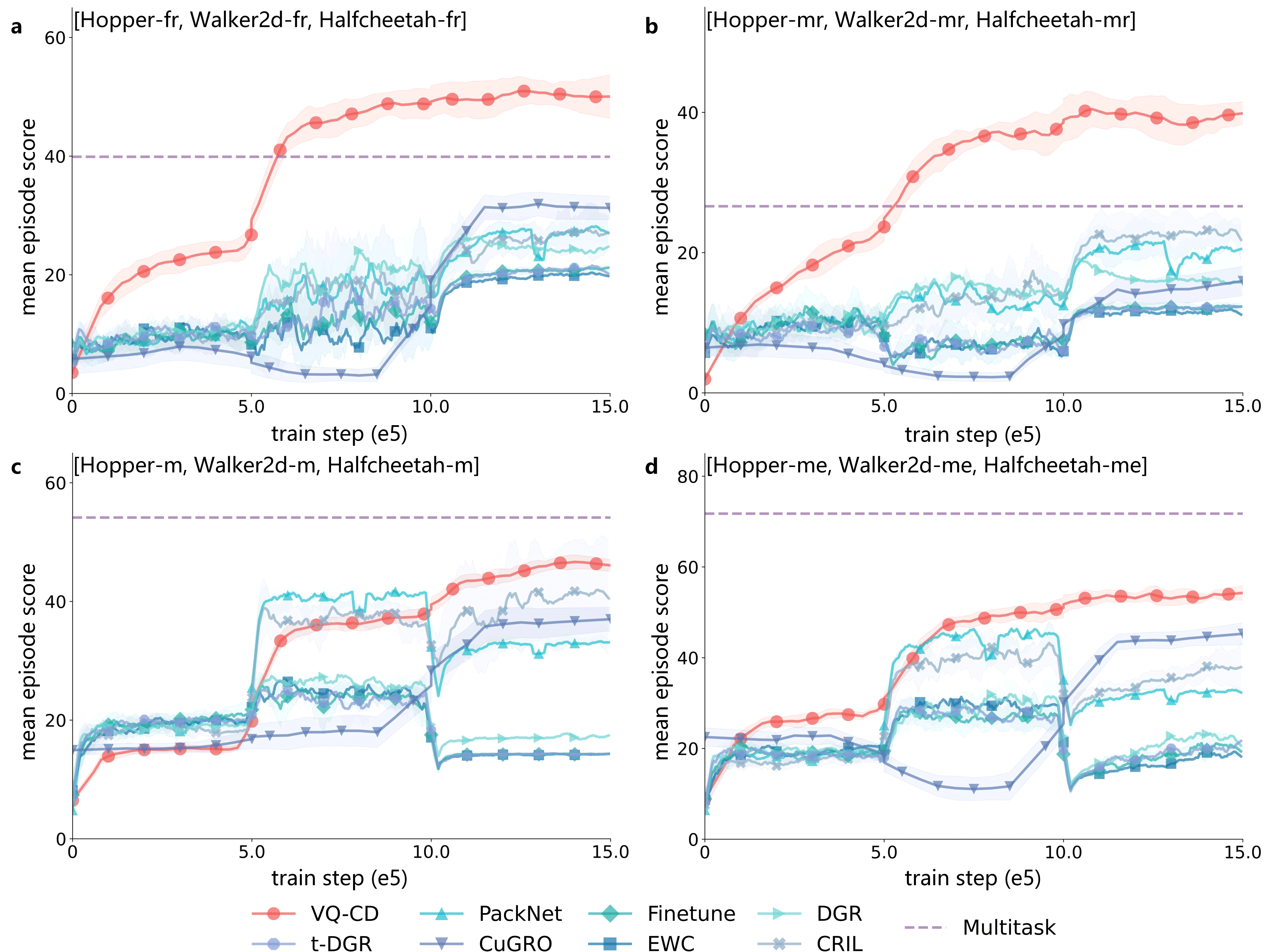}}
 \caption{The comparison on the arbitrary CL settings. We select the D4RL tasks to formulate the CL task sequence. We leverage state and action padding to align the spaces. The experiments are conducted on various dataset qualities, where the results show that our method surpasses the baselines not only at the expert datasets but also at the non-expert datasets, which illustrates the wide task applicability of our method. The datasets characteristic ``fr'', ``mr'', ``m'', and ``me'' represent ``full-replay'', ``medium-replay'', ``medium'', and ``medium-expert'', respectively. ``Hopper", ``Walker2d", and ``Halfcheetah" are the different environments.}
 \label{offline d4rl comparison on main body}
 \end{center}
 \vspace{-0.3cm}
 \end{figure*}

The arbitrary CL settings correspond to the second question we want to answer: \emph{Can we use the proposed space alignment method to enable \ourmodel{} to adapt to incoming tasks with various spaces?}

To answer the above question, we select D4RL to formulate the CL task sequence because of the various state and
action spaces, and the results are shown in Figure~\ref{offline d4rl comparison on main body}.
Considering the dataset qualities of D4RL~\citep{fu2020d4rl}, we choose different dataset quality settings and report the mean episode score that is calculated with $\frac{R_i-R_{random}}{R_{expert}-R_{random}}*100$.
Generally, from the four sub-experiments (\textbf{a}, \textbf{b}, \textbf{c}, and \textbf{d}), we can see that our method (\ourmodel{}) surpasses these baselines in all CL settings.
Especially in the CL settings (Figure~\ref{offline d4rl comparison on main body} \textbf{a} and \textbf{b}), where the datasets contain low-quality trajectories, our method achieves a large performance margin even compared with the Multitask method.
We can attribute the reason to the return-based action generation that helps our method distinguish different quality trajectories and generate high-reward actions during evaluation, as well as the selective weights activation that can reserve the previous knowledge and reduce forgetting.
While other methods just possess the ability to continue learning and lack the ability to separate different qualities and actions, thus leading to poor performance.
For trajectory qualities that are similar across the datasets (Figure~\ref{offline d4rl comparison on main body} \textbf{c} and \textbf{d}), we can see lower improvement gains between our method and baselines.
However, it should be noted that our method can still reach better performance than other baselines.
Apart from the padding alignment, we also conduct experiments (Figure~\ref{offline vq_d4rl comparison}) on baselines that adopt our pre-trained QSA module to align state and action spaces in Appendix~\ref{Experiments of Baselines Equipped QSA}.

\subsection{Ablation Study}

\begin{figure*}[t!]
 \begin{center}
\ifthenelse{\equal{\figureresolution}{low resolution}}
    {\includegraphics[angle=0,width=0.99\textwidth]{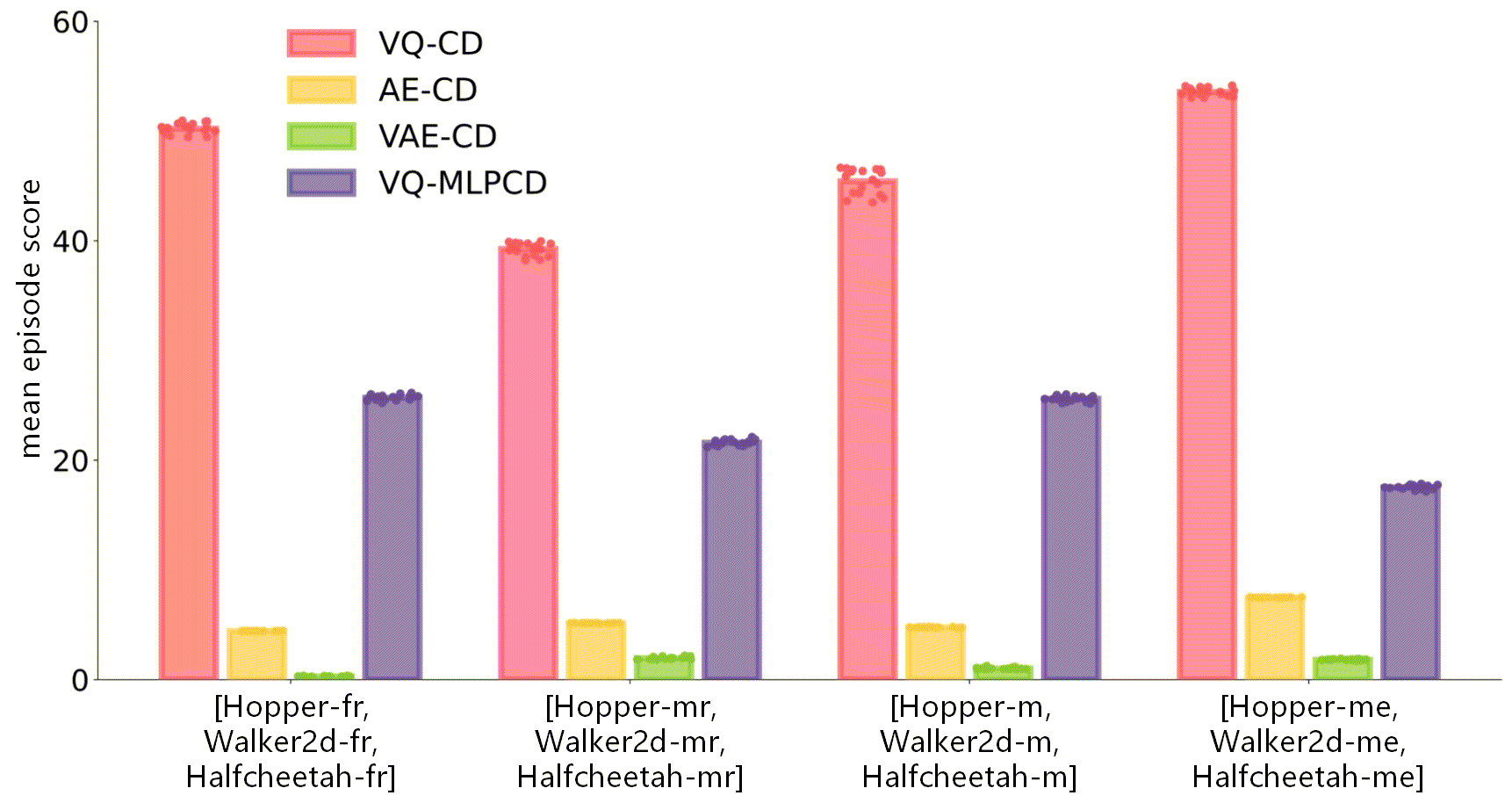}}
    {\includegraphics[angle=0,width=0.99\textwidth]{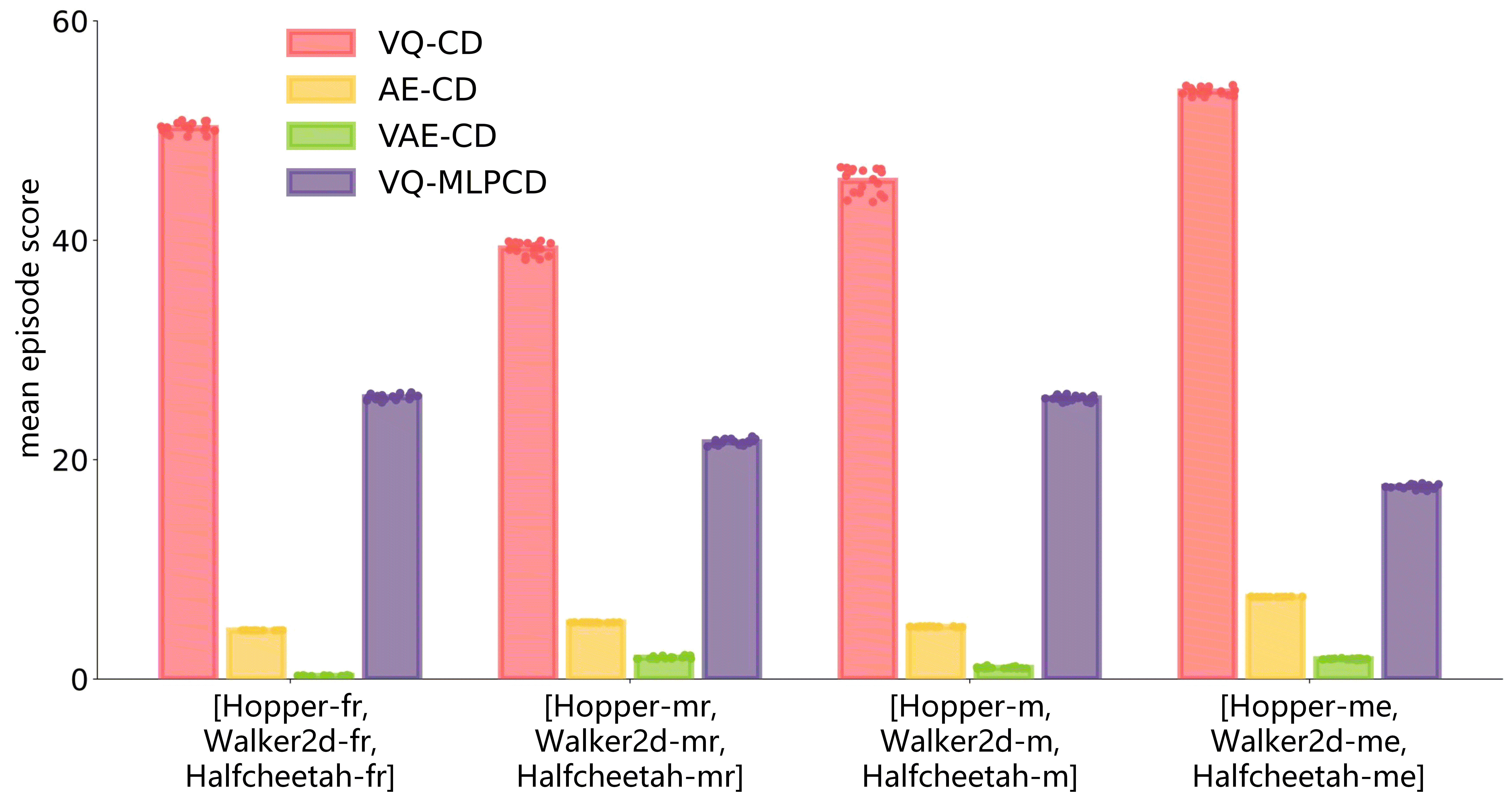}}
 \caption{The ablation study of space alignment module and diffusion network structure. For each type of ablation study, we fix the other same and retrain the model on four D4RL CL settings.}
 \label{d4rl ablation study}
 \end{center}
 \end{figure*}

In this section, we want to investigate the influence of different modules of \ourmodel{}.
Thus, the experiments contain two investigation directions: space alignment module ablation study and diffuser network structure ablation study.
To show the importance of vector quantization, we change the space alignment module with auto-encoder (AE) and variational auto-encoder (VAE).
Based on this modification, we retrain our method and report the results in Figure~\ref{d4rl ablation study}.
The results show significant improvements in the D4RL CL settings, illustrating the importance and effectiveness of vector quantization in our method.
Compared with AE-CD, VAE-CD performs poorer on all D4RL CL settings. 
The reason lies in that the implicit Gaussian constraint on each dimension may hurt the space alignment.
Compared with the codebook in \ourmodel{}, AE-CD may cause a bigger difference between aligned features produced by AE (shown in Table~\ref{state and action difference comparison}), posing challenges for the diffusion model to model the distribution of the aligned features and leading to low performance.
As for the diffuser network structure, we conduct the selective weights activation on the mlp-based and unet-based diffusion models.
The latter structure is beneficial to making decisions with temporal information inside the trajectories, leading to higher performance evaluation.

\begin{table*}[t!]
\centering
\small
\caption{The feature difference between the aligned features produced by the space alignment module. We randomly sample thousands of aligned state and action features to calculate the difference.}
\label{state and action difference comparison}
\resizebox{0.99\textwidth}{!}{
\begin{tabular}{l | r | r | r | r}
\toprule
\specialrule{0em}{1.5pt}{1.5pt}
\toprule
Method & \multicolumn{2}{c|}{\ourmodel{}} & \multicolumn{2}{c}{AE-CD}\\
\midrule[1pt]
feature difference & state difference & action difference & state difference & action difference \\
\midrule[1pt]
$[\text{Hopper-fr,Walker2d-fr,Halfcheetah-fr}]$ & 8.83\tiny{$\pm$1.98} & 4.54\tiny{$\pm$0.74} & 51.31\tiny{$\pm$26.91}& 14.06\tiny{$\pm$2.09} \\
$[\text{Hopper-mr,Walker2d-mr,Halfcheetah-mr}]$ & 9.03\tiny{$\pm$1.97} & 4.45\tiny{$\pm$0.74} & 48.12\tiny{$\pm$21.94} & 15.39\tiny{$\pm$3.71} \\
$[\text{Hopper-m,Walker2d-m,Halfcheetah-m}]$ & 8.53\tiny{$\pm$1.56} & 4.22\tiny{$\pm$0.79} & 42.27\tiny{$\pm$24.29} & 13.59\tiny{$\pm$2.63} \\
$[\text{Hopper-me,Walker2d-me,Halfcheetah-me}]$ & 8.93\tiny{$\pm$2.00} & 4.05\tiny{$\pm$0.56} & 57.91\tiny{$\pm$36.94} & 13.93\tiny{$\pm$3.20} \\
\bottomrule
\specialrule{0em}{1.5pt}{1.5pt}
\bottomrule
\end{tabular}}
\vspace{-0.1cm}
\end{table*}

 \begin{figure*}[t!]
 \begin{center}
\ifthenelse{\equal{\figureresolution}{low resolution}}
    {\includegraphics[angle=0,width=0.99\textwidth]{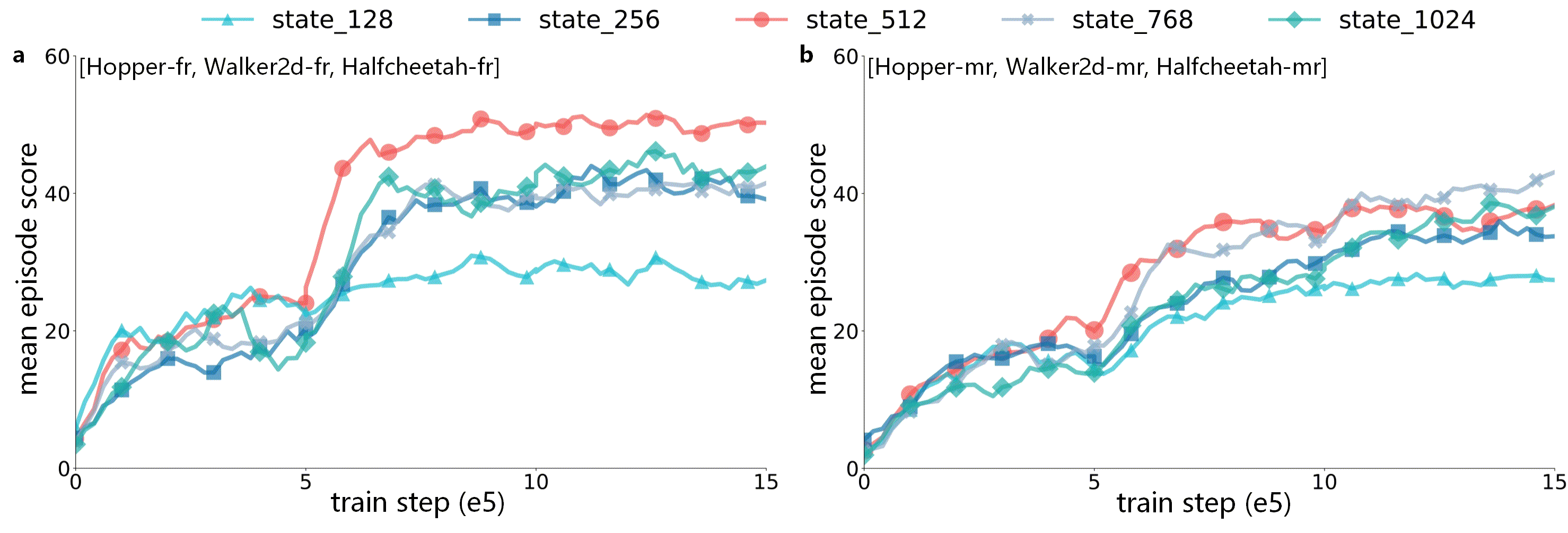}}
    {\includegraphics[angle=0,width=0.99\textwidth]{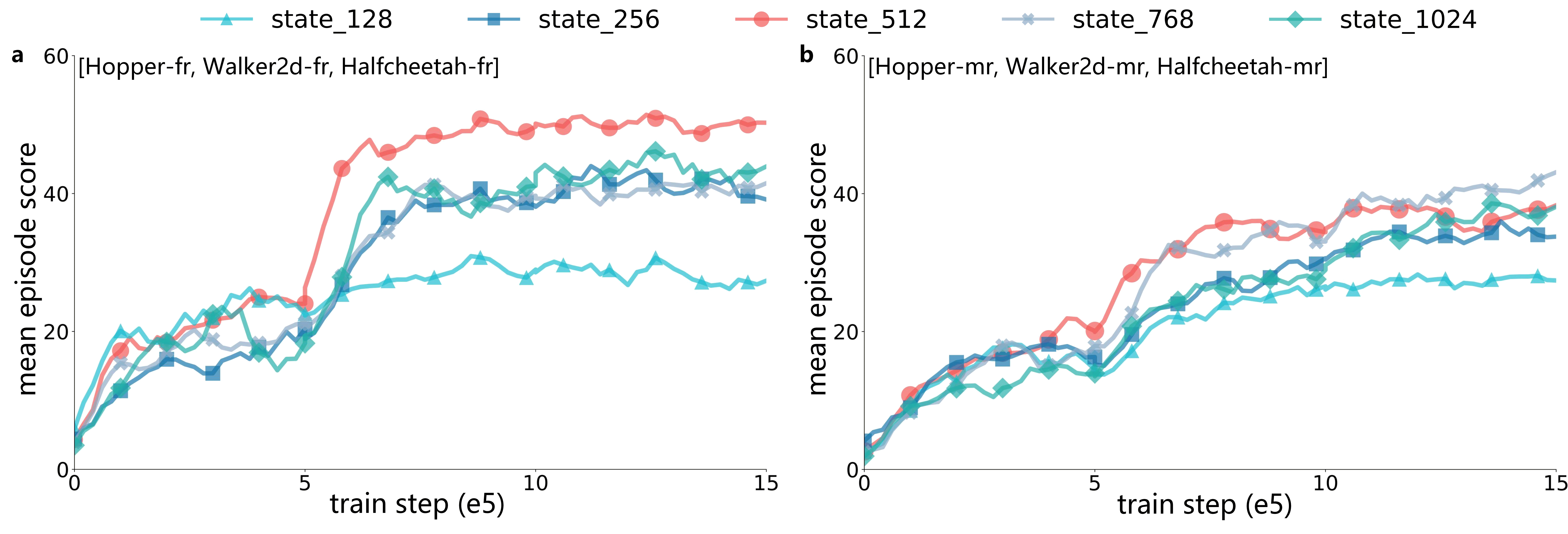}}
 \caption{The effects of different codebook sizes about the states.}
 \label{parameter sensitivity on state}
 \end{center}
 \vspace{-0.3cm}
 \end{figure*}

 \begin{figure*}[t!]
\hspace{-1em}
 \begin{center}
\ifthenelse{\equal{\figureresolution}{low resolution}}
    {\includegraphics[angle=0,width=0.99\textwidth]{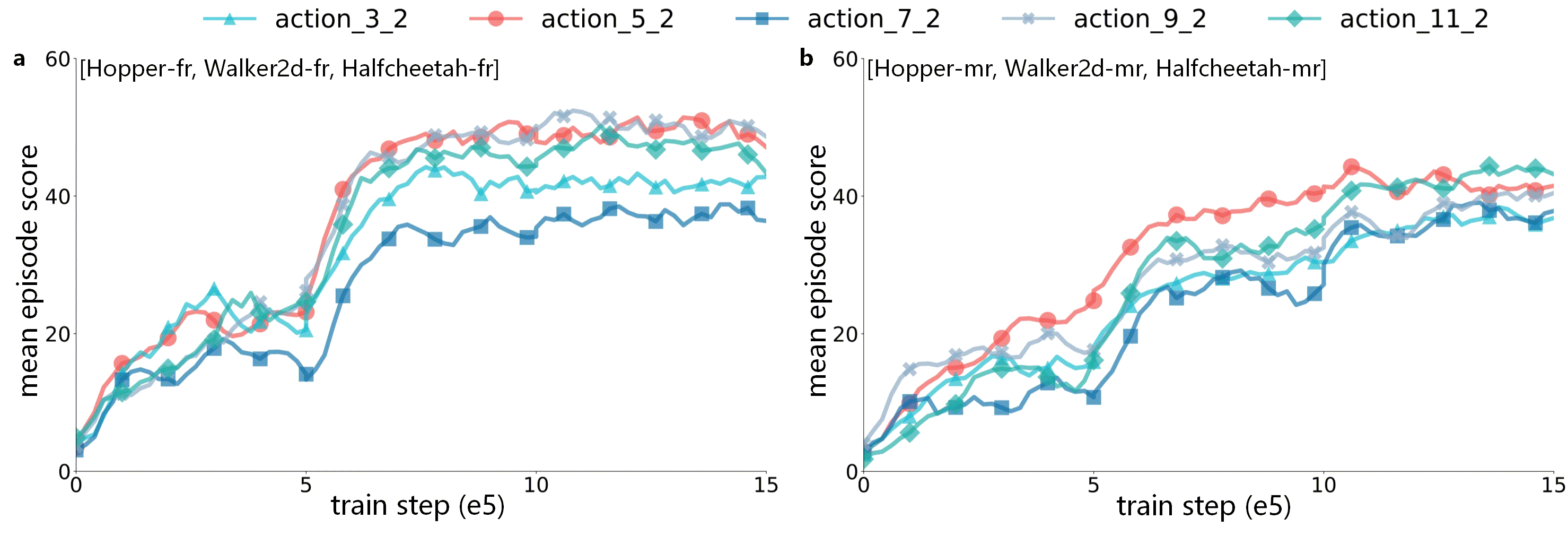}}
    {\includegraphics[angle=0,width=0.99\textwidth]{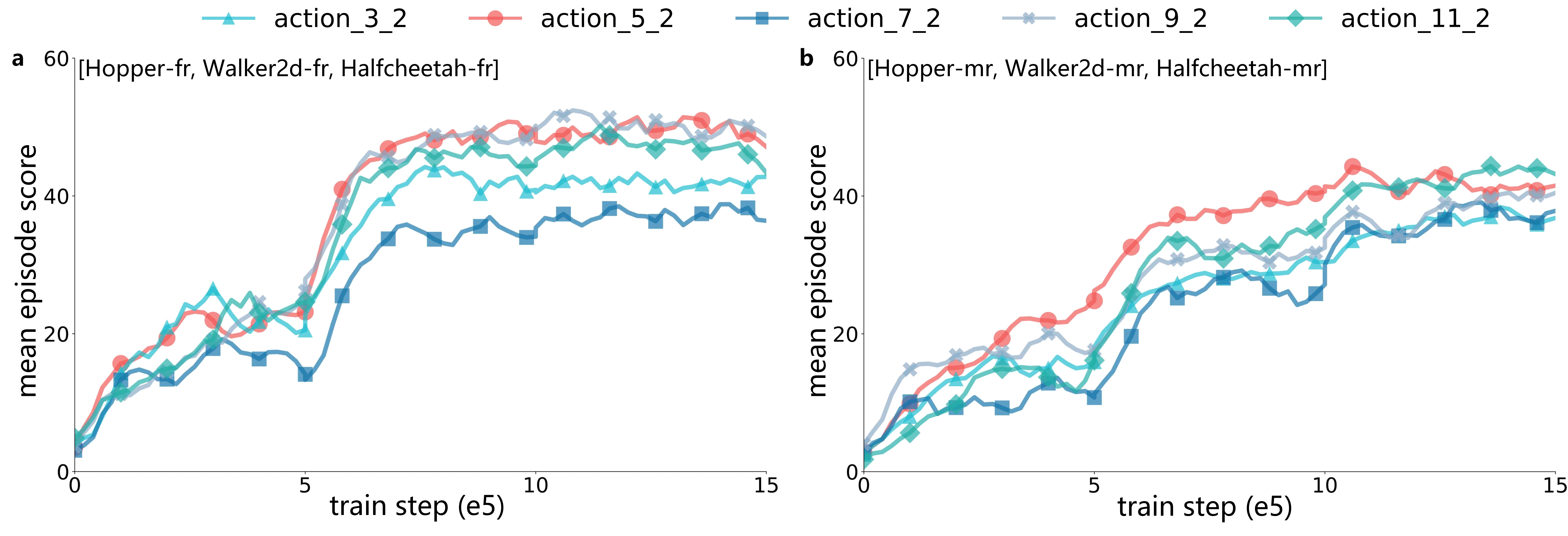}}
 \caption{The effects of the number of latent vectors about the actions.}
 \label{parameter sensitivity on action}
 \end{center}
 \vspace{-0.3cm}
 \end{figure*}

\subsection{Parameter Sensitivity Analysis}\label{Parameter Sensitivity Analysis}

When performing on the aligned feature with diffusion models, the hyperparameters of state and action of the quantized spaces alignment module matter.
Usually, the complexity of states is more significant than the actions, so the codebook size controls the performance of reconstruction.
Thus, we investigate the effect of different codebook sizes and report the results in Figure~\ref{parameter sensitivity on state}.
Obviously, a small codebook size limits performance, and a negative effect arises when it exceeds a certain value, such as 512. 
As for the actions, we believe the actions can be decomposed into several small latent vectors, and the number of latent vectors is crucial for reconstructing actions.
Similarly, we also see the same trend in Figure~\ref{parameter sensitivity on action}, which inspires us that more latent vectors are not always better.

\section{Conclusion}

In this paper, we propose Vector-Quantized Continual Diffuser, called \ourmodel{}, which opens the door to training on any CL task sequences.
The advantage of this general ability to adapt to any CL task sequences stems from the two sections of our framework: the selective weights activation diffuser (SWA) module and the quantized spaces alignment (QSA) module.
SWA preserves the previous knowledge by separating task-related parameters with task-related masking.
QSA aligns the different state and action spaces so that we can perform training in the same aligned space.
Finally, we show the superiority of our method by conducting extensive experiments, including conventional CL task settings (identical state and action spaces) and general CL task settings (various state and action spaces).
The results illustrate that our method achieves the SOTA performance by comparing with 16 baselines on 15 continual learning task settings.

\clearpage
\bibliography{reference}
\bibliographystyle{iclr2025_conference}

\clearpage
\appendix
\large
\begin{center}
   \emph{Appendix of ``Solving Continual Offline RL through Selective Weights Activation on Aligned Spaces''} 
\end{center} 
\normalsize

\section{Algorithm}

\subsection{Pseudocode of \ourmodel{}}\label{appendix of pseudocode}

\begin{algorithm}
\caption{Vector-Quantized Continual Diffuser (\ourmodel{})}
\label{algorithm}
\LinesNumbered
\KwIn{Noise prediction model $\epsilon_{\theta}$, state and action quantized model $f_q(\theta_e,\theta_d,\theta_q)$, tasks set $\mathcal{M}_i, i\in\{1,...,I\}$, each task training step $\Omega$, max diffusion step $K$, sequence length $T_e$, state dimension $d_s$, action dimension $d_a$, reply buffer $D_i, i\in\{1,...,I\}$, noise schedule $\alpha_{0:K}$ and $\beta_{0:K}$}
\KwOut{$\epsilon_{\theta}$, $f_q(\theta_e,\theta_d,\theta_q)$}
\textbf{Initialization:} $\theta$, $\theta_e$, $\theta_d$, and $\theta_q$ \\
Separate the state-action trajectories of $D_i, i\in\{1,...,I\}$ into state-action sequences with length $T_e$ \\
// \textbf{Quantized Spaces Alignment (QSA) Pretraining} \\
\For{each task $i$}
{
   \For{each train epoch}
   {
      \For{each train epoch}
      {
         Sample states and actions from task $i$'s buffer $D_i$ \\
         Calculate the quantization loss and reconstruction loss \\
         Updating the parameters of $\theta_e$, $\theta_d$, and $\theta_q$ by solving the problem of Equation~\ref{QSA problem} \\
      }
   }
   Save the pre-trained QSA module \\
}
// \textbf{Selective Weights Activation (SWA) Diffuser Training} \\
\For{each task $i$}
{
   Generate the task-related mask $M_i$ for task $i$\\
   \For{each train epoch}
   {
       \For{each train step $m$}
       {
           Sample $b$ sequences $\tau_i^{0}=\{s_{i,t:t+T_e}, a_{i,t:t+T_e}\}\in\mathbb{R}^{b\times T_e\times (d_s+d_a)}$ from task $i$'s buffer $D_i$ \\
           Obtain the quantized state and action feature $s_{z_q}, a_{z_q}$ with the QSA module \\
           Formulate $s_{z_q}, a_{z_q}$ as sequences $\tau_{i,z_q}^{0}=\{s_{i,z_q,t:t+T_e}, a_{i,z_q,t:t+T_e}\}$\\
           Sample the corresponding discounted returns $R_{i,t:t+T_e}$ from task $i$'s buffer $D_i$ \\
           Sample diffusion time step $k\sim \text{Uniform}(K)$ \\
           Sample Gaussian noise $\epsilon \sim \mathcal{N}(0,\bm{I}), \epsilon\in\mathbb{R}^{b\times T_e\times (d_{s_{z_q}}+d_{a_{z_q}})}$ \\
           Obtain $\tau_{i,z_q}^{k}$ by adding noise to $\tau_{i,z_q}^{0}$ \\
           Perform the forward propagation with Equation~\ref{forward propagation masking} \\
           Train $\epsilon_{\theta}$, $f_{return}(\phi)$, and $f_{time}(\varphi)$ according to Equation~\ref{diffusion model train loss} \\ 
       }
   }
   Save task $i$'s related models as $\epsilon_{i*\Omega, \theta}$ \\
}
// \textbf{Weights Assembling} \\
Construct new models $\Tilde{\epsilon}_{\theta}$ with the same structure as $\epsilon_{ \theta}$ \\
\For{each task $i$}
{
   Extract the task-related parameters $W_i$ with mask information $M_i$ from $\epsilon_{i*\Omega, \theta}$ \\
   Fill the corresponding task-related parameters $W_i=M_i\circ W_i$ into $\Tilde{\epsilon}_{\theta}$ \\
}
\end{algorithm}

The training of \ourmodel{} (Pseudocode is shown in Algorithm~\ref{algorithm}) contains three stages.
1) We first pre-train the QSA module for space alignment, as shown in lines 4-13, where we mainly want to solve the constrained problem of Equation~\ref{QSA problem}.
2) Then, in lines 15-31, for each task $i$, we generate the task-related mask $M_i$ followed by a standard diffusion model training process (Refer to Equation~\ref{diffusion model train loss} and Equation~\ref{forward propagation masking} for the training loss) on the aligned state and action spaces.
3) Finally, we assemble the task-related weights $W_i$ together with the mask information $\{M_i|i\in[1:I]\}$ according to $W=\sum M_i\circ W[i*\Omega]$, where $\Omega$ is the training steps for each CL task, and $W[i*\Omega]$ is the weights checkpoints of $\epsilon_{i*\Omega, \theta}$.
It is noted that the pre-training of the QSA module and the training of the SWA module can be merged together, i.e., for each task $i$, we can first train the QSA module related to task $i$ and then train the SWA module.

\subsection{Hyperparameters}\label{hyperparameters}
We classify the hyperparameters into three categories: QSA module-related, SWA module-related, and training-related hyperparameters.
We use the learning rate schedule when pre-training the QSA module, so the VQ learning rate decreases from 1e-3 to 1e-4.
In our experiments, the maximum diffusion steps are set as 200, and the default structure is Unet.
Usually, it is time-consuming for the diffusion-based model to generate actions in RL.
Thus, we consider the speed-up technique DDIM~\citep{song2020denoising} and realize it in our method to improve the generation efficiency during evaluation.
For all models, we use the Adam~\citep{kingma2014adam} optimizer to perform parameter updating.

\begin{table}[t!]
\centering
\caption{The hyperparameters of \ourmodel{}.}
\label{our method hyper}
\begin{tabular}{l | l l}
\toprule
 & Hyperparameter & Value\\
 \midrule
 \multirow{12}{*}{QSA section}   & network backbone & MLP\\
                               & hidden dimension of QSA module & 256 \\ 
                               & commitment cost coefficient & 0.25 \\
                               & codebook embedding limit $\rho$ & 3.0 \\
                               & state codebook size per task & 512 \\ 
                               & number of state latent & 10 \\
                               & state latent dimension & 2 \\
                               & action codebook size per task & 512 \\ 
                               & number of action latent & 5 \\
                               & action latent dimension & 2 \\
                               & alignment type & VQ/AE/VAE \\
                               & VQ learning rate & [1e-4,1e-3] \\
\midrule
\multirow{10}{*}{SWA section}   & network backbone & Unet/MLP\\
                               & hidden dimension & 256 \\
                               & sequence length $T_e$  & 8\\
                               & diffusion learning rate & 3e-4\\
                               & guidance value & 0.95 \\
                               & mask rate & $1/I$ \\
                               & condition dropout & 0.25 \\
                               & max diffusion step $K$ & 200 \\
                               & sampling speed-up stride & 20 \\
                               & condition guidance $\omega$ & 1.2 \\
                               & sampling type of diffusion & DDIM \\      
\midrule
\multirow{4}{*}{Training} & loss function & MSE \\
                           & batch size & 32 \\
                           & optimizer & Adam \\
                           & discount factor $\gamma$ & 0.99 \\
\bottomrule
\end{tabular}
\end{table}

\subsection{Computation}\label{computation}
We conduct the experiments on NVIDIA GeForce RTX 3090 GPUs and NVIDIA A10 GPUs, and the CPU type is Intel(R) Xeon(R) Gold 6230 CPU @ 2.10GHz. 
Each run of the experiments spanned about 24-72 hours, depending on the algorithm and the length of task sequences.

\begin{table*}[h!]
\centering
\small
\caption{The comparison of generation speed with different generation steps under the CL setting of Ant-dir task-4-18-26-34-42-49. In the main body of our manuscript, we use the 10 diffusion steps setting for all experiments.}
\label{Efficiency Analysis of Generation Speed}
\resizebox{0.99\textwidth}{!}{
\begin{tabular}{l | r  r  r  r  r  r}
\toprule
\specialrule{0em}{1.5pt}{1.5pt}
\toprule
Diffusion steps & 200 (original) & 100 & 50 & 25 & 20 & 10 \\
\midrule
sampling speed-up stride & 1 (original) & 2 & 4 & 8 & 10 & 20 \\
\midrule
\makecell[l]{Time consumption of \\ per generation (s)} & 5.73\tiny{$\pm$0.29} & 2.88\tiny{$\pm$0.21} & 1.41\tiny{$\pm$0.16} & 0.71\tiny{$\pm$0.18} & 0.58\tiny{$\pm$0.17} & 0.29\tiny{$\pm$0.15} \\
\midrule
Speed-up ratio & 1× & 1.99× & 4.06× & 8.07× & 9.88× & 19.76× \\
\bottomrule
\specialrule{0em}{1.5pt}{1.5pt}
\bottomrule
\end{tabular}}
\end{table*}

\begin{table*}[t!]
\centering
\small
\caption{The GPU memory consumption. }
\label{The GPU memory consumption}
\resizebox{0.85\textwidth}{!}{
\begin{tabular}{l | l | r }
\toprule
\specialrule{0em}{1.5pt}{1.5pt}
\toprule
domain & CL task setting & GPU memory consumption (GB)\\
\midrule
 \multirow{4}{*}{D4RL} & $[\text{Hopper-fr,Walker2d-fr,Halfcheetah-fr}]$  & 4.583 \\
 & $[\text{Hopper-mr,Walker2d-mr,Halfcheetah-mr}]$ & 4.583 \\
 & $[\text{Hopper-m,Walker2d-m,Halfcheetah-m}]$ & 4.583 \\
 & $[\text{Hopper-me,Walker2d-me,Halfcheetah-me}]$ & 4.583 \\
 \midrule
 \multirow{2}{*}{Ant-dir} & task-10-15-19-25 & 4.711\\ 
 & task-4-18-26-34-42-49 & 5.955\\
 \midrule
 CW & CW10 & 5.897\\
\bottomrule
\specialrule{0em}{1.5pt}{1.5pt}
\bottomrule
\end{tabular}}
\end{table*}

\subsection{Generation Speed-up Technique}\label{Generation Speed-up Technique}
The time and memory consumption of diffusion models is attributed to the mechanism of diffusion generation process that requires multiple computation rounds to generate data~\cite{ho2020denoising}.
Fortunately, previous studies provide useful speed-up strategies to accelerate the generation process~\citep{nichol2021improved, song2020denoising}.
In this paper, we adopt DDIM as the default generation speed-up technique and reduce the reverse diffusion generation step to 10 compared to the original 200 generation steps.
In Table~\ref{Efficiency Analysis of Generation Speed}, we use the CL setting of Ant-dir task-4-18-26-34-42-49 as an example to compare the time consumption of different generation steps.
Compared with the original 200 diffusion steps, we can see that incorporating DDIM will significantly (\textbf{19.76×}) improve the efficiency of generation. 
In the experiments, we find that 10 diffusion steps setting performs well on performance and generation efficiency.
Thus, we set the default sampling speed-up stride to 20, and the diffusion step is 200/20=10 steps.

\subsection{Memory Consumption}\label{Memory Consumption}

In Table~\ref{The GPU memory consumption}, we report the GPU memory consumption during the training process.
We mainly consider the experiments on the D4RL, Ant-dir, and CW CL tasks.
We can change the first block of the diffusion model to make our model suitable for a longer CL task sequence.
For example, we expand the dimension length from 512 to 1024 when switching the CL training task from `task-10-15-19-25' to `task-4-18-26-34-42-49'.

\section{Additional Experiments}

\subsection{QSA Module Loss Analysis}

\begin{figure*}[t!]
 \begin{center}
\ifthenelse{\equal{\figureresolution}{low resolution}}
    {\includegraphics[angle=0,width=0.99\textwidth]{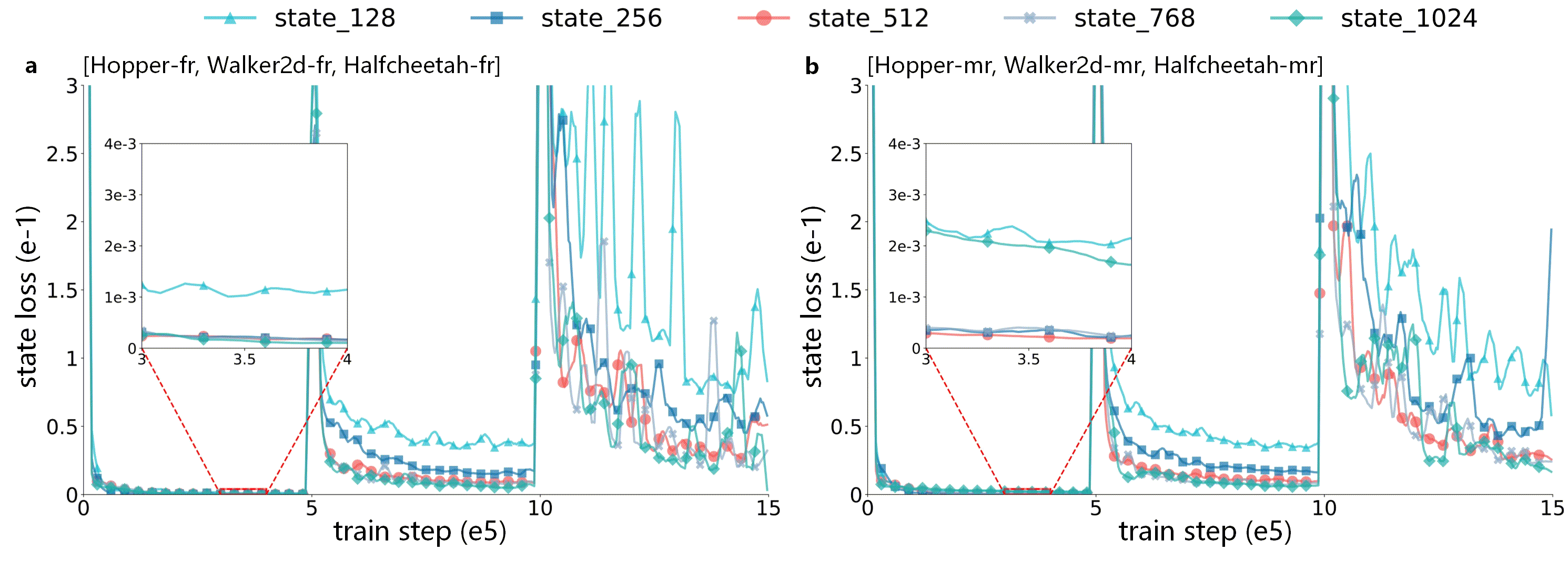}}
    {\includegraphics[angle=0,width=0.99\textwidth]{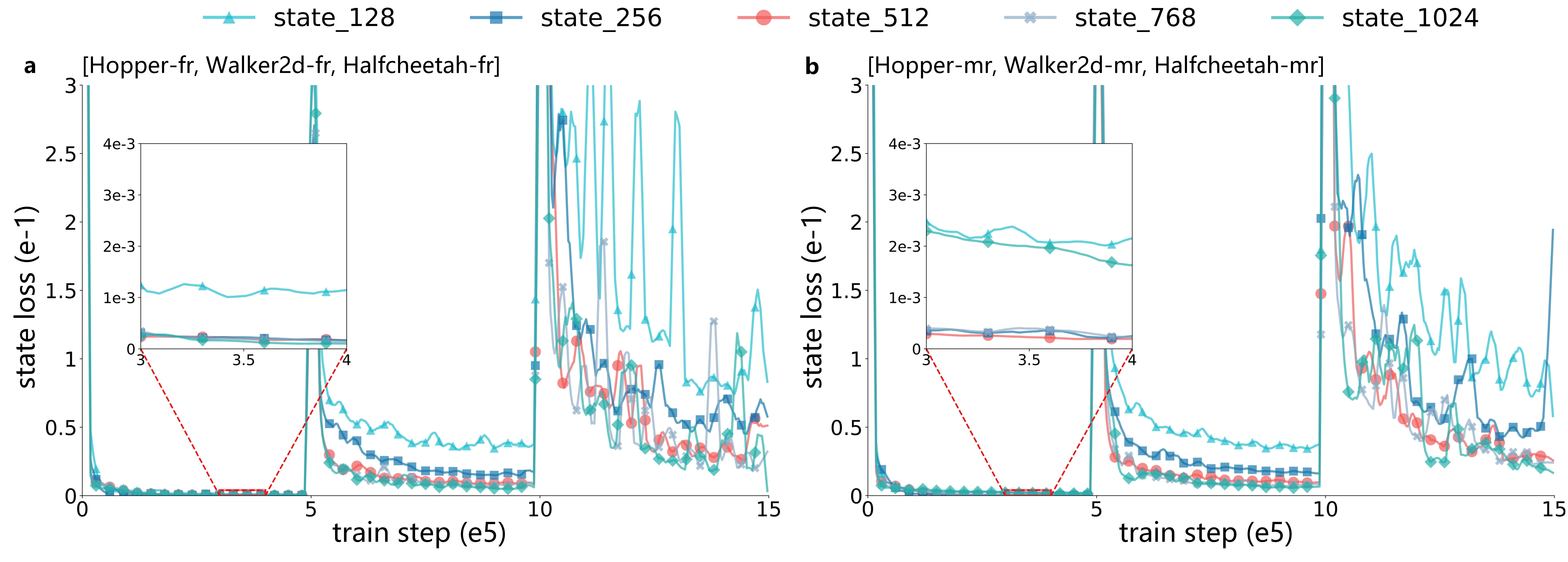}}
 \caption{The QSA module loss under different codebook sizes about states. We explore five codebook size settings: 128, 256, 512, 768, and 1024. The red line represents the experimental codebook size setting for states.}
 \label{QSA module loss of state}
 \end{center}
 \end{figure*}

Under the same hyperparameter settings in Section~\ref{Parameter Sensitivity Analysis}, we report the loss of the QSA Module to investigate the effects of codebook size and latent number.
For the states, we investigate the influence of codebook size, where we set codebook size as 128, 256, 512, 768, 1024, and select D4RL CL setting [Hopper-fr, Walker2d-fr, Halfcheetah-fr] and [Hopper-mr, Walker2d-mr, Halfcheetah-mr] as the example.
The results are shown in Figure~\ref{QSA module loss of state}, where we train the QSA module on each task for 5e5 steps.
We can see that for states, a codebook size of 512 is good enough for aligning the different tasks' state spaces.
A larger codebook size, such as 768 and 1024 in Figure~\ref{QSA module loss of state} \textbf{a} and \textbf{b}, will not bring significant loss improvements.
Smaller codebook sizes can not provide sufficient latent vectors to map the state spaces to a uniform space.

\begin{figure*}[t!]
 \begin{center}
\ifthenelse{\equal{\figureresolution}{low resolution}}
    {\includegraphics[angle=0,width=0.99\textwidth]{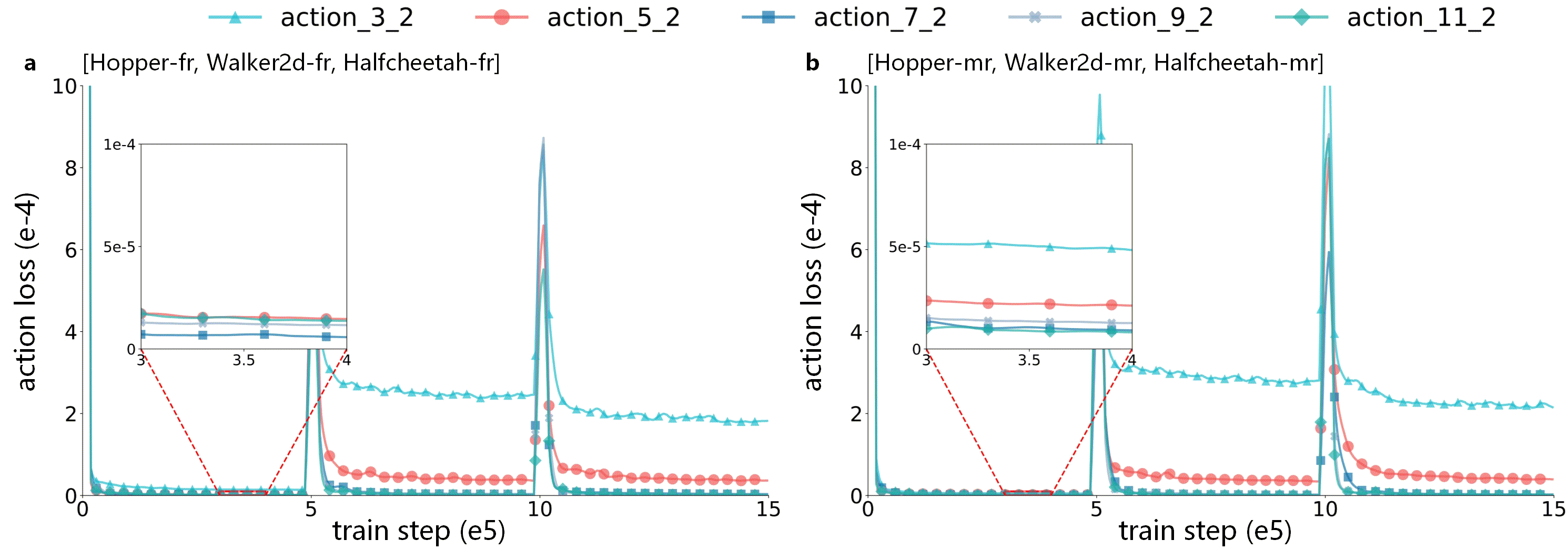}}
    {\includegraphics[angle=0,width=0.99\textwidth]{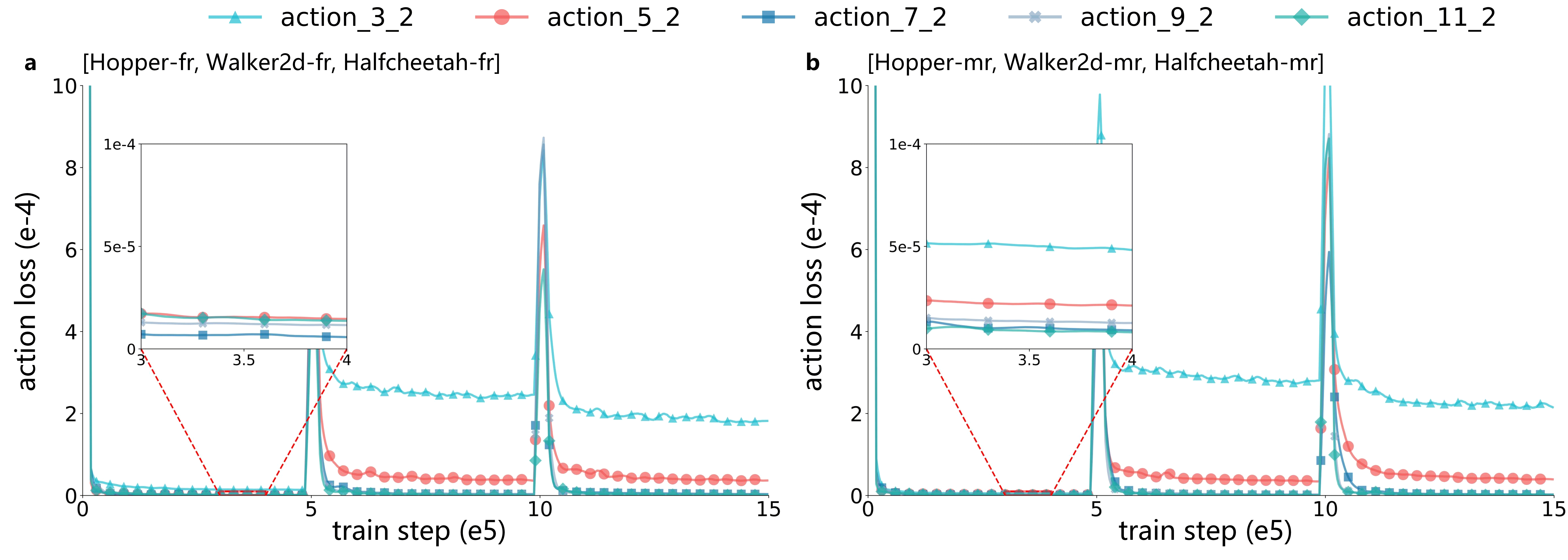}}
 \caption{The QSA module loss under different latent numbers about actions. The setting includes 3, 5, 7, 9, and 11, which correspond to the aligned action space sizes 6, 10, 14, 18, and 22. The red line represents the experimental latent numbers setting for actions.}
 \label{QSA module loss of action}
 \end{center}
 \end{figure*}

For the action, we select the latent number to explore the QSA action loss and report the results in Figure~\ref{QSA module loss of action}.
We can see the same trend that has been seen in QSA state loss (Figure~\ref{QSA module loss of state}).
Though the lower loss value of the more latent number indicates that we should use more action latent vectors, we find that the gap between action latent number settings 5 and 7 is small when we increase computation resources.
Besides, we also see inconspicuous performance gains in the final performance in Figure~\ref{parameter sensitivity on action}, which urges us to use 5 as the default action latent number setting.
For the action latent vector dimension, we directly use 2 as the default setting.

\subsection{Experiments of Task Order Shuffling}\label{Experiments of Task Order Shuffling}
\begin{figure*}[t!]
 \begin{center}
\ifthenelse{\equal{\figureresolution}{low resolution}}
    {\includegraphics[angle=0,width=0.99\textwidth]{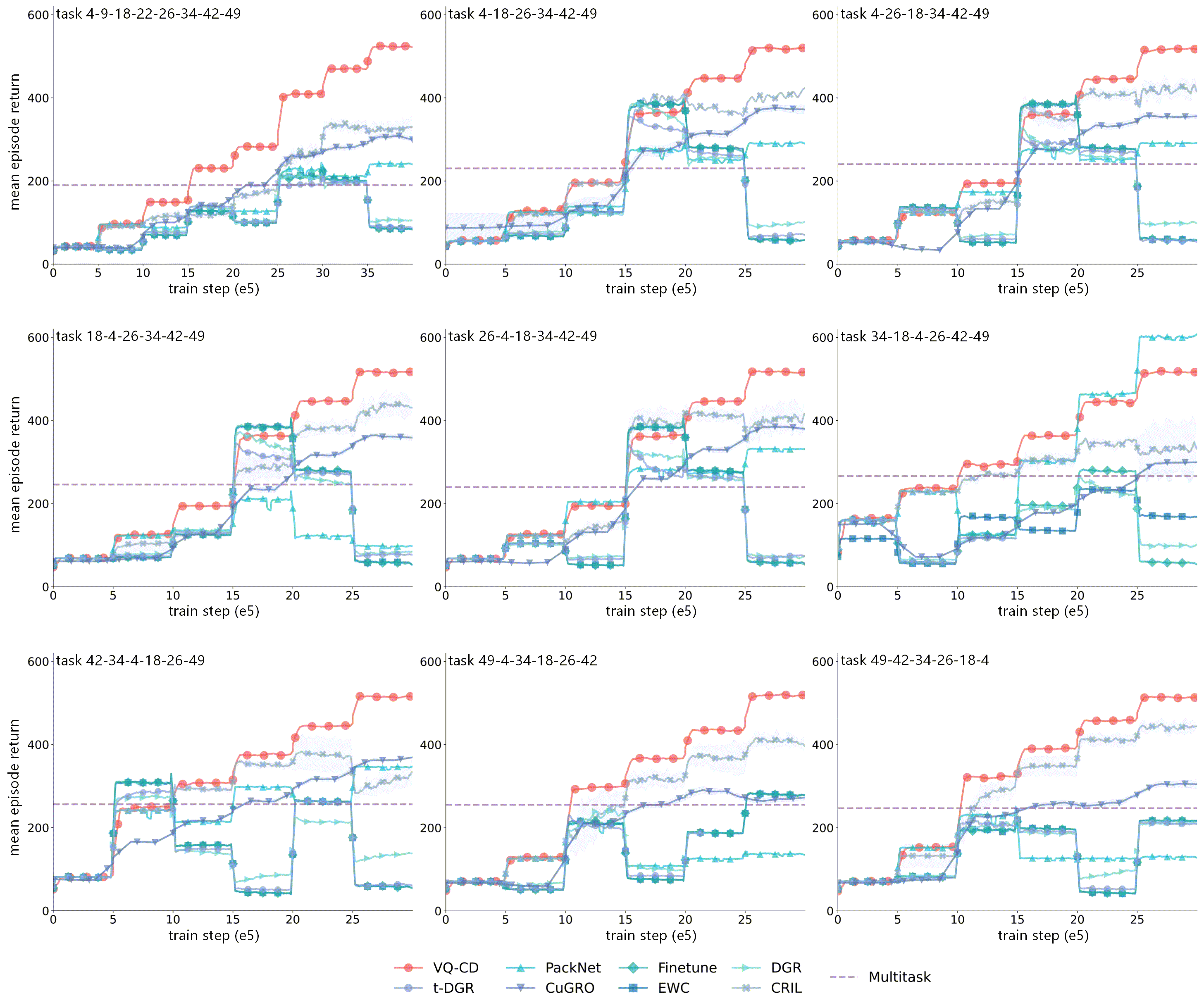}}
    {\includegraphics[angle=0,width=0.99\textwidth]{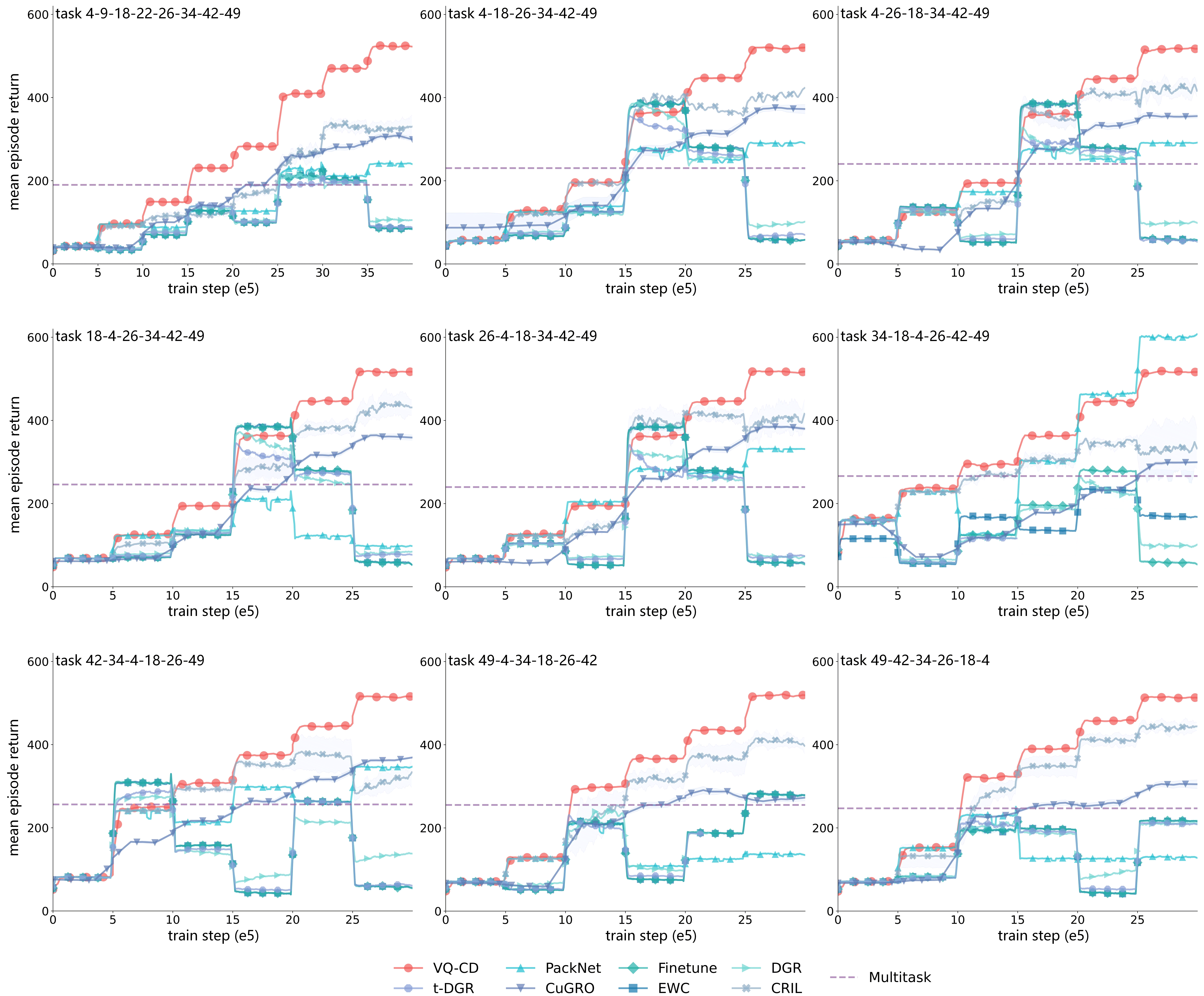}}
 \caption{The experiments of Ant-dir with shuffled task order. We investigate the influence of shuffled task order in the Ant-dir environment, where the experiments include inserting new tasks into the predefined task order `4-18-26-34-42-49' and disrupting the tasks order.}
 \label{antdir perturb_task_sequence}
 \end{center}
 \end{figure*}

To investigate the influence of task order in CORL, we choose Ant-dir as the testbed and change the task order for new CL training.
We change the task order by inserting new tasks into the predefined task order `4-18-26-34-42-49' and disrupting the task order.
We can see from the results shown in Figure~\ref{antdir perturb_task_sequence} that our method achieves the best performance in almost all CL task orders.
The task order will affect the final performance of other baselines. 
For example, CRIL performs better in the task orders `task 18-4-26-34-42-49' and `task 49-42-34-26-18-4' than in other task order experiments.
Another example is PackNet, which achieves the best performance only in the task order `task 34-18-4-26-42-49'.
Different from the baselines, whose performance fluctuates with the changing of task orders, our method (\ourmodel{}) shows stable training performance no matter what task orders are defined.

\subsection{Experiments of Baselines Equipped QSA}\label{Experiments of Baselines Equipped QSA}
\begin{figure*}[t!]
\vspace{-1em}
 \begin{center}
\ifthenelse{\equal{\figureresolution}{low resolution}}
    {\includegraphics[angle=0,width=0.99\textwidth]{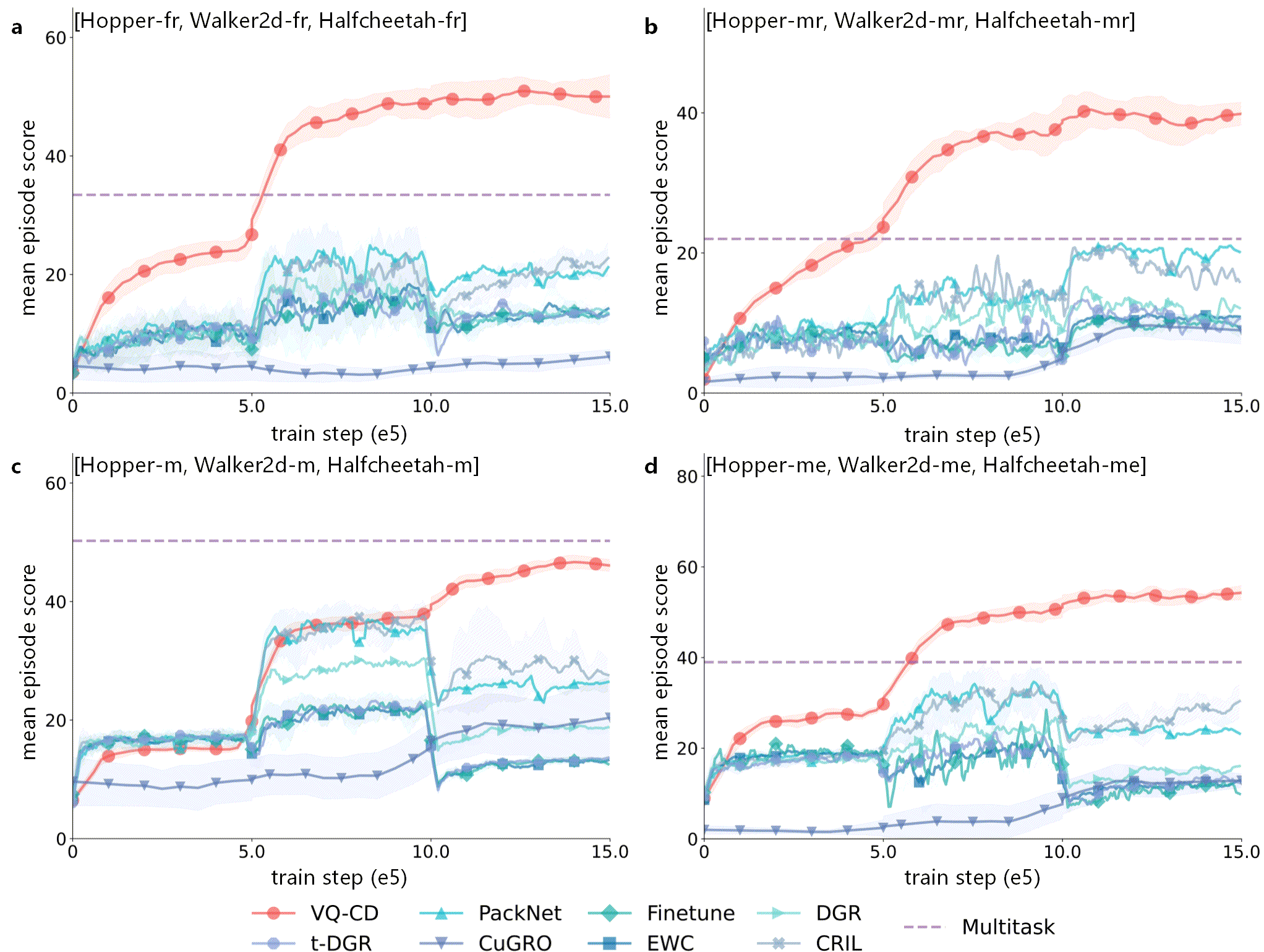}}
    {\includegraphics[angle=0,width=0.99\textwidth]{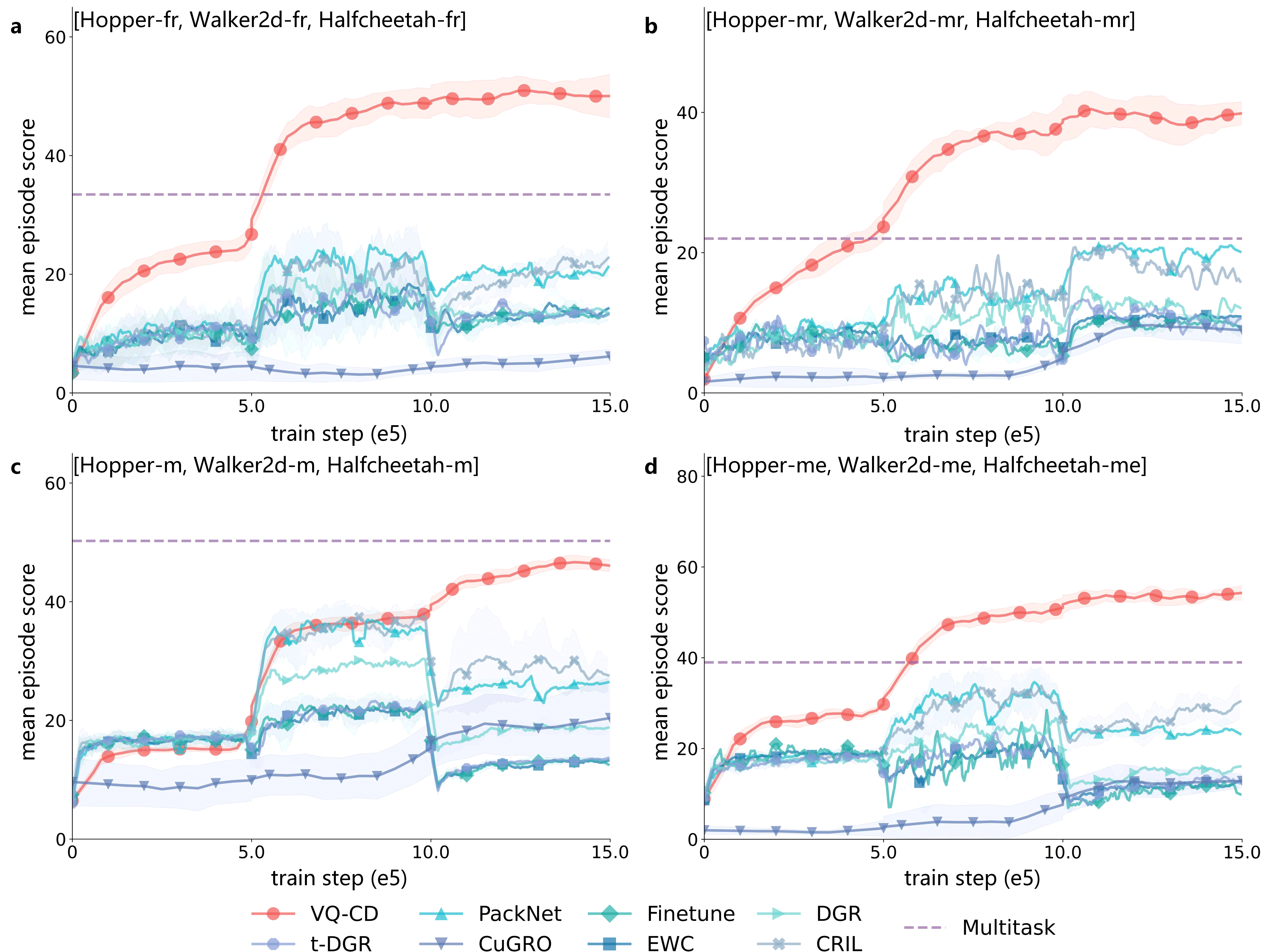}}
 \caption{The comparison on the arbitrary CL settings. We select the D4RL tasks to formulate the CL task sequence.
 In order to align the state and action spaces, we use the pre-trained QSA module (the same as our method) to provide aligned spaces during training. The experiments are conducted on various dataset qualities, where the results show that our method surpasses the baselines not only at the expert datasets but also at the non-expert datasets, which illustrates the wide task applicability of our method. The datasets characteristic ``fr'', ``mr'', ``m'', and ``me'' represent ``full-replay'', ``medium-replay'', ``medium'', and ``medium-expert'', respectively. ``Hopper", ``Walker2d", and ``Halfcheetah" are the different environments.}
 \label{offline vq_d4rl comparison}
 \end{center}
 \vspace{-0.3cm}
 \end{figure*}

In Section~\ref{Experimental Results}, we report the comparison of our method and baselines in the arbitrary CL settings, where in the D4RL CL settings, we adopt state and action padding to align the state and action spaces.
Apart from the state and action padding, we can also use the pre-trained QSA module to align the different state and action spaces.
In Figure~\ref{offline vq_d4rl comparison}, we report the results of baselines equipped with QSA.
When introducing the QSA, the model is actually trained on the feature space rather than the original state and action spaces, which makes it hard to learn for these baselines proposed from the traditional CL setting.
From the results, we can also see that our method still achieves the best performance compared with these baselines.
Considering the results of Figure~\ref{d4rl ablation study} (VQ-MLPCD) and Figure~\ref{offline vq_d4rl comparison} (VQ baselines), we can see the importance of complementary sections: QSA and SWA.

\begin{table*}[t!]
\centering
\small
\caption{The comparison of time consumption per update between sparse and dense (normal) optimizers. We compare these two types of optimizers on the CL settings and find that when we first use the normal optimizer, such as Adam, to train the model and then use weights assembling to obtain the final model, the total physical time consumption is significantly smaller than sparse optimizer (e.g., sparse Adam).}
\label{The optimizer time consumption}
\resizebox{0.99\textwidth}{!}{
\begin{tabular}{l | l | r r}
\toprule
\specialrule{0em}{1.5pt}{1.5pt}
\toprule
\multirow{2}{*}{domain} & \multirow{2}{*}{CL task setting} & \multicolumn{2}{c}{time consumption per update (s)}\\
\cline{3-4}
 &  & dense optimizer & sparse optimizer\\
\midrule
 \multirow{4}{*}{D4RL} & $[\text{Hopper-fr,Walker2d-fr,Halfcheetah-fr}]$  & 0.089\tiny{$\pm$0.219} & 0.198\tiny{$\pm$0.224}\\
 & $[\text{Hopper-mr,Walker2d-mr,Halfcheetah-mr}]$ & 0.096\tiny{$\pm$0.223} & 0.197\tiny{$\pm$0.223} \\
 & $[\text{Hopper-m,Walker2d-m,Halfcheetah-m}]$ & 0.089\tiny{$\pm$0.211} & 0.195\tiny{$\pm$0.224} \\
 & $[\text{Hopper-me,Walker2d-me,Halfcheetah-me}]$ & 0.090\tiny{$\pm$0.223} & 0.206\tiny{$\pm$0.225} \\
 \midrule
 \multirow{2}{*}{Ant-dir} & task-10-15-19-25 & 0.062\tiny{$\pm$0.064} & 0.239\tiny{$\pm$0.282}\\ 
 & task-4-18-26-34-42-49 & 0.064\tiny{$\pm$0.061} & 0.214\tiny{$\pm$0.270}\\
 \midrule
 CW & CW10 & 0.061\tiny{$\pm$0.065} & 0.218\tiny{$\pm$0.286}\\
\bottomrule
\specialrule{0em}{1.5pt}{1.5pt}
\bottomrule
\end{tabular}}
\end{table*}

\subsection{Supporting Tasks Training Beyond the Pre-defined Task Sequence}\label{Supporting Tasks Training Beyond the Pre-defined Task Sequence}
\begin{figure*}[t!]
\vspace{-1em}
 \begin{center}
\ifthenelse{\equal{\figureresolution}{low resolution}}
    {\includegraphics[angle=0,width=0.99\textwidth]{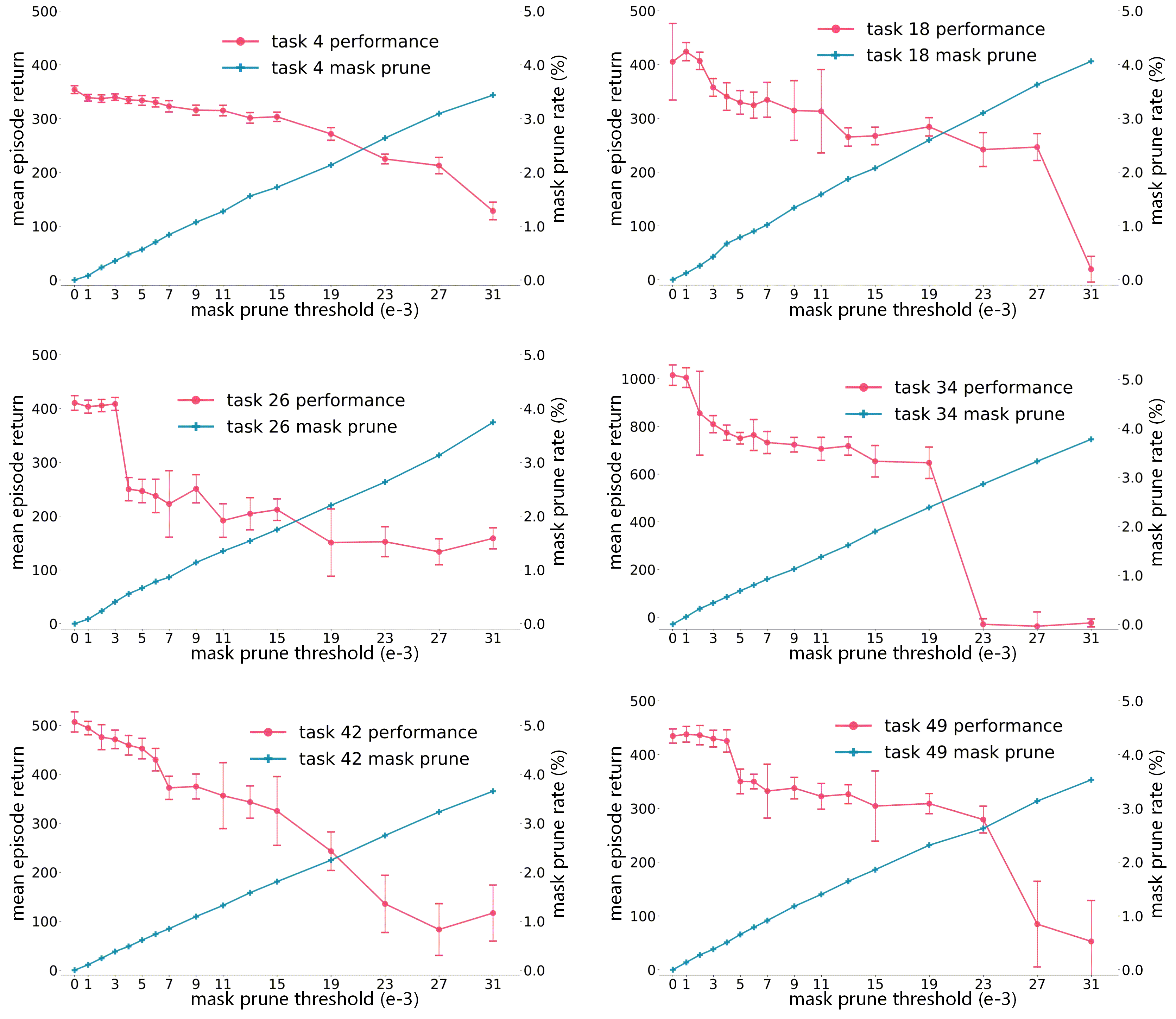}}
    {\includegraphics[angle=0,width=0.99\textwidth]{figures/ant_dir_mask_prun.png}}
 \caption{The mask pruning experiments of Ant-dir `task 4-18-26-34-42-49'. We investigate the task pruning according to the absolute weight values, i.e., we release the weights to train on potential new tasks according to the mask prune threshold.}
 \label{offline ant_dir mask prune}
 \end{center}
 \vspace{-0.3cm}
 \end{figure*}

After training on pre-defined task sequences, we may hope the model has the capacity to support training on potential tasks, which means that we need more weights or weight masks. 
Releasing weight masks that are used to learn previous tasks is a straightforward choice when the total weights are fixed.
We conduct the experiments of mask pruning on Ant-dir `task 4-18-26-34-42-49' and report the performance and weight mask prune rate when pruning weight masks according to certain absolute value thresholds in Figure~\ref{offline ant_dir mask prune}. 
The results illustrate that we can indeed release some weight masks under the constraint of preserving 90\% or more performance compared with the unpruned model.
On the other hand, we can also see that this mask pruning method can only provide finite capacity for tasks beyond the pre-defined task sequence.
We postpone the systematic investigation of mask pruning to future works.

\subsection{Time Consumption of Different Optimizers}\label{Time Consumption of Different Optimizers}

In the CL settings of our experiments, we compare two types of optimizers and find that when we first use the normal optimizer, such as Adam, to train the model and then use weights assembling to obtain the final model, the total physical time consumption is significantly smaller than sparse optimizer (e.g., sparse Adam).
Thus, we propose the weights assembling to obtain the final well-trained model after the training rather than suffering huge time burden of sparse optimizer during the training.

\end{document}